\documentclass[review]{elsarticle}  	

\usepackage{graphics}		
\usepackage{epsfig}			
\usepackage{amsmath}		
\usepackage{amssymb}		
\usepackage{hyperref}
\usepackage{algorithm}		
\usepackage{algpseudocode}	
\usepackage{array}
\usepackage{float}
\usepackage{booktabs}
\usepackage{bm}
\usepackage{subcaption}

\usepackage{epstopdf}
\renewcommand{\vec}[1]{\bm{#1}}
\newcommand{\mat}[1]{\bm{#1}}
\newcolumntype{C}[1]{>{\centering\let\newline\\\arraybackslash\hspace{0pt}}m{#1}}

\begin{document}
\begin{frontmatter}

\title{\LARGE \bf Flexible Human-Robot Cooperation Models for Assisted Shop-floor Tasks}

\author{Kourosh Darvish}
\author{Francesco Wanderlingh}
\author{Barbara Bruno\fnref{mark:corresponding}}
\fntext[mark:corresponding]{Corresponding author. \textit{Email address}: \texttt{barbara.bruno@unige.it}}
\author{Enrico~Simetti}
\author{Fulvio Mastrogiovanni}
\author{Giuseppe Casalino}

\address{Department of Informatics, Bioengineering, Robotics and Systems Engineering,\\ University of Genoa, via Opera Pia 13, 16145 Genoa, Italy.}

\begin{abstract}

The Industry 4.0 paradigm emphasizes the crucial benefits that collaborative robots, i.e., robots able to work alongside and together with humans, could bring to the whole production process. In this context, an enabling technology yet unreached is the design of flexible robots able to deal at all levels with humans' intrinsic variability, which is not only a necessary element for a comfortable working experience for the person, but also a precious capability for efficiently dealing with unexpected events. In this paper, a sensing, representation, planning and control architecture for flexible human-robot cooperation, referred to as FlexHRC, is proposed. FlexHRC relies on wearable sensors for human action recognition, AND/OR graphs for the representation of and reasoning upon cooperation models, and a Task Priority framework to decouple action planning from robot motion planning and control.

\end{abstract}

\begin{keyword}
Human-Robot Cooperation; Smart Factory; AND/OR graph; Task Priority control; Wearable Sensing.
\end{keyword}

\end{frontmatter}


\section{Introduction}
\label{sec:Introduction}

According to the Industry 4.0 paradigm, manufacturing is expected to undergo an important paradigm shift involving the nature of shop-floor environments. 
One of the main ideas put forth in smart factories is \textit{getting closer} to customers, increasing their satisfaction through a high degree of personalization and just in time goods delivery.
This poses serious challenges to shop-floor operators, insofar as work stress, fatigue and eventually alienation are concerned, with repercussions also on work quality and faulty work pieces.

Among the recommendations to reduce such drawbacks on human operators, collaborative robots have been proposed to work alongside humans to perform a series of tasks traditionally considered stressful, tiring or difficult \cite{Lenz2011}.
Clearly, this proposal implies a number of challenges related to human-robot interaction both at the physical and cognitive level of the cooperation \cite{DeSantis2008, Hayes2016, Alami2017}, which depend also on their type \cite{Helms2002}.
Beside basic safety considerations, which are a necessary prerequisite \cite{KuehnHaddadin2017}, a number of key issues must be taken into account: sensing and human activity recognition \cite{Pedrocchietal2013}, definition of suitable cooperation models to reach certain goals \cite{Shah2011, Kock2011, Johannsmeier2017}, robot action planning and execution in the presence of humans \cite{Alami2017}, just to name a few.

When focusing on human-robot cooperation, these considerations can be synthesized in four functional specifications focusing on improving the operator's \textit{user experience} \cite{Meneweger2015}:
\begin{enumerate}
\item[$F_1$] Operators should not be forced to follow a strict, predefined sequence of operations, but should be allowed to decide what actions to perform \textit{on the fly}, subject to their adherence to the overall cooperation goals. 
\item[$F_2$] Robots should trade-off between providing operators with optimal suggestions about next actions to perform and reacting appropriately when operators do not follow such instructions.
\item[$F_3$] Robots should decouple action planning (meaningful for operators) and motion planning and control, also when the workspace is partially unknown.
\item[$F_4$] Operators should not be required to limit their motions, e.g., to stay in front of a collaborative robot, to have their actions duly monitored.
\end{enumerate}

In this paper, a sensing, representation, planning and control architecture for flexible human-robot cooperation, referred to as FlexHRC, is proposed. FlexHRC deals with the above specifications by design, in particular enforcing flexibility at two levels:
\begin{enumerate}
\item[$R_1$] Although robots suggest actions to perform based on \textit{optimality} considerations and the goal to achieve, operators can choose an action without following the robot's suggestions \cite{Hawkins2014}, while the robot reacts and plans for the next action accordingly \cite{Bertenthal1996, Vesper2010, loehr2013}.   
\item[$R_2$] Although robot operations are well-defined in terms of motion trajectories, reactive behaviors allow for dealing with partially unknown or dynamic workspaces, e.g., to perform obstacle avoidance, without the need for trajectory re-planning \cite{Srivastavaetal2014, Simetti2016d}.
\end{enumerate}

To this aim, FlexHRC implements a hybrid, reactive-deliberative human-robot cooperation architecture for \textit{assisted cooperation} \cite{Helms2002, Kruger2009} integrating different modules, namely:
(i) human action recognition using \textit{wearable sensors}, which does not pose any constraint on operator motions \cite{Bruno2013};
(ii) representation of human-robot cooperation models and reasoning using \textit{AND/OR graphs} \cite{Sandersonetal1988, Hawkins2014, Johannsmeier2017};
(iii) control schemes based on a \textit{Task Priority} framework to decouple human/robot action planning from robot motion planning and control \cite{Simetti2016d}.

The paper is organized as follows.
Section \ref{sec:background} discusses related work.
Cooperation models and the associated sensing, reasoning and robot motion processes are described in Section \ref{sec:architecture}.
Experimental results are presented and discussed in Section \ref{sec:evaluation}.
Conclusions follow.

\section{Background}
\label{sec:background}

During the past few years, human-robot interaction gained much attention in the research literature.
Whilst approaches focused on cooperation consider aspects related to natural interaction with robots, e.g., targeting human-robot coordination in joint action \cite{Sebanz2006, Valdesolo2010, Huber2013}, this analysis focuses on the human-robot cooperation process from the perspective of the functional specifications discussed above.

The problem of allowing humans and robots to perform open-ended cooperation by means of coordinated activity ($F_1$) did not receive adequate attention so far.
An approach highlighting the challenge is presented in \cite{Shah2011}, in which an execution planning and monitoring module adopts two teamwork modes, i.e., when humans and robots are equal partners and when humans act as leaders. 
On the one hand, a reference shared plan is generated off line, and actions are allocated to a human or a robot according to their capabilities. 
On the other hand, coordination is achieved by an explicit step-by-step, speech-based, human to robot communication, which makes user experience cumbersome and unnatural in most cases. 

The ability of robots to mediate between high-level planning and low-level reactive behaviors has been subject of huge debates in the past three decades. 
When it comes to human-robot cooperation, the need arises to balance the requirements of reaching a well-defined goal (e.g., a joint assembly) and providing human co-workers with as much freedom as possible ($F_2$).
A number of conceptual elements for joint and coordinated operations are identified in \cite{Vesper2010}.
The authors propose a \textit{minimalistic} architecture to deal with aspects related to agents cooperation.
In particular, a formalism to define goals, tasks and their representation, as well as the required monitoring and prediction processes, is described.
The work discussed in \cite{Alami2017} significantly extends the notions introduced in \cite{Vesper2010} to focus on \textit{social} human-robot interaction aspects. 
The architecture makes an explicit use of \textit{symbol anchoring} to reason about human actions and cooperation states.
An approach sharing some similarities with FlexHRC is described in \cite{Johannsmeier2017}.
As in the proposed approach, AND/OR graphs are used to sequence actions for the cooperation process. 
However, unlike FlexHRC, action sequences cannot be switched at runtime, but are determined off line in order to optimize graph-based metrics.   
As a matter of fact, the possibility of multiple cooperation models is provided for, although off line: optimal paths on the AND/OR graph are converted to \textit{fixed} action sequences, and then executed without any possible variation.
In a similar way, multiple cooperation models are considered in \cite{Hawkins2014}, where an AND/OR graph is converted to a nondeterministic finite state machine for representation, and later to a probabilistic graphical model for predicting and monitoring human actions, as well as their timing.

The development of sensing and control architectures able to integrate and coordinate action planning with motion planning and control is an active research topic.
However, the challenge is typically addressed to deal with cases where planning cannot be guaranteed to be \textit{monotone}, i.e., when sensory information must be used to validate the plan during execution \cite{Agrawaletal2016}.  
Its application to human-robot cooperation tasks ($F_3$) has not been fully addressed in the literature. 
An approach in that direction is described in \cite{Toussaint2016}, where an integrated approach to Monte Carlo based action planning and trajectory planning via Programming by Demonstration is adopted in a scenario of toolbox assembly. 
Concurrent activities are formalized using a Markov decision process, which determines when initiating and terminating each human or robot action.

Finally, a few approaches consider the issue of allowing human operators to retain a certain freedom of motion or action when interacting with a robot ($F_4$), but at the price of introducing a few assumptions in the process \cite{BAUER2008, Cartmill2011}.
A Bayesian framework is used in \cite{Huber2013} to track a human hand position in the workspace with the aim of predicting an action's time-to-completion.
The hand must be clearly visible for the estimate to be accurate, which limits certain motions with the aim of obtaining information about time-related aspects of the cooperation.
The opposite approach is adopted in \cite{Shah2011}, where an extended freedom of motion is obtained resorting to speech-based communication to indicate performed actions to the robot, as well as action start and end times.
The obvious drawback of this approach relies on the fact that such a communication act must be voluntary, and therefore human stress and fatigue may jeopardize the will to do it.
A more comprehensive approach is described in \cite{Alami2017}, which integrates human body position (determined by an external sensory system, e.g., motion capture), deictic gestures, gaze and verbal communication to determine a number of human actions.
An \textit{external} system for human activity recognition may be of difficult deployment in a shop-floor environment, and occlusions may occur nonetheless.

From this focused analysis, it emerges that although a number of approaches have been discussed, which take the above specifications into account, they do so only partially.
FlexHRC attempts to provide a holistic and integrated solution to these heterogeneous challenges.

\begin{figure}[!t]
\centering
\includegraphics[width=0.9\columnwidth]{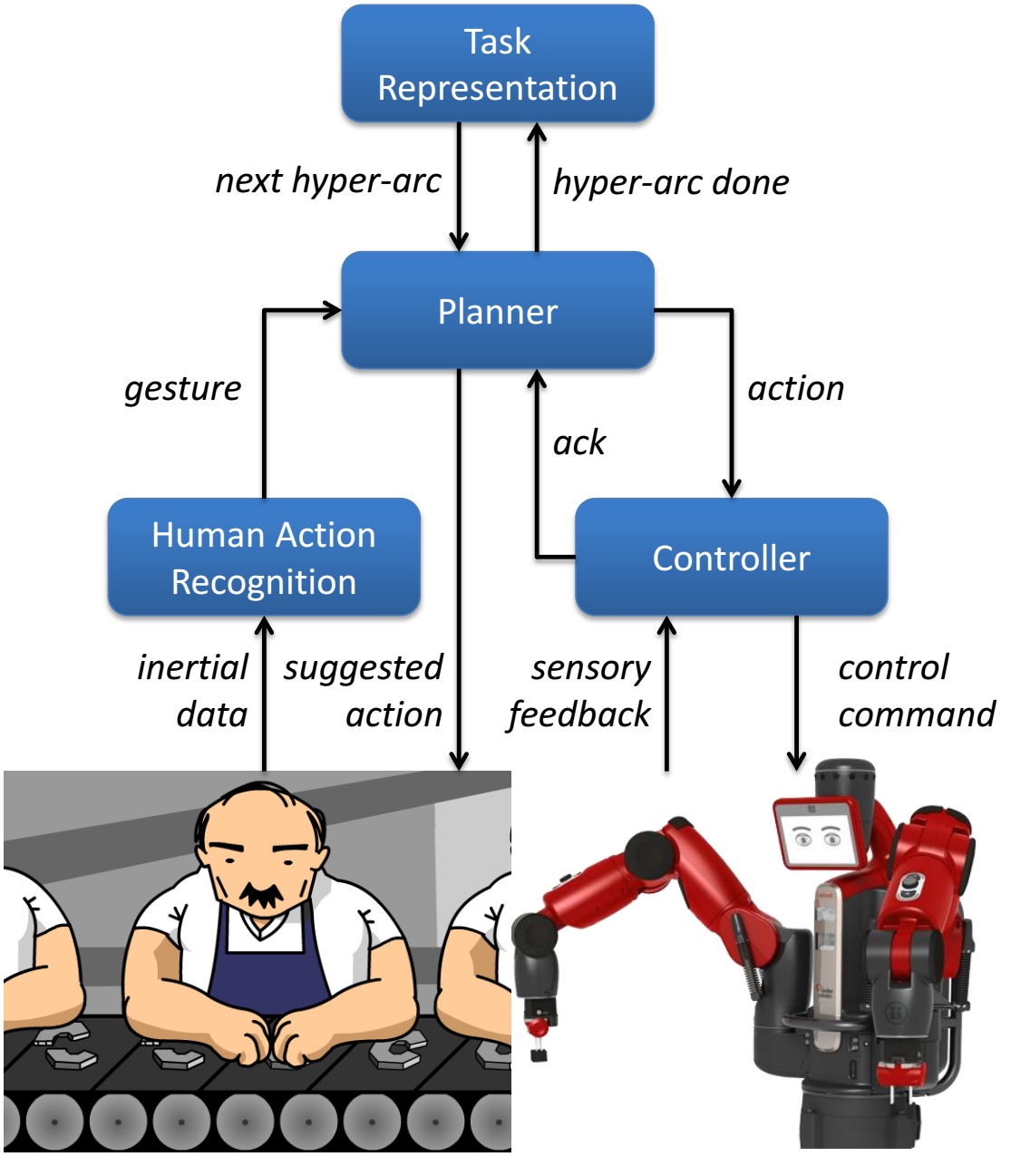}
\caption{The FlexHRC system's architecture: modules and data flow.}
\label{fig:SysArchitecture}
\end{figure}

\section{An Architecture for Flexible Human-Robot Cooperation}
\label{sec:architecture}

FlexHRC is based on a distributed hybrid reactive-deliberative architecture (Figure \ref{fig:SysArchitecture}).
Four modules are involved.
\textit{Task Representation} maintains cooperation models for the activities to be carried out, and interacts with the \textit{Planner} to decide which action a human or a robot should execute next.
These two modules are based on AND/OR graphs and an \textit{ad hoc} online graph traversal procedure.
To this aim, the \textit{Planner} is informed by human gestures via the \textit{Human Action Recognition} module and by robot actions via the \textit{Controller}. 
Adopting the approach described in \cite{Bruno2013,Bruno2014}, \textit{Human Action Recognition} retrieves data from wearable devices and classifies inertial data streams according to a number of predefined \textit{gesture models}.
Finally, the \textit{Controller} is tasked with motion control and execution, and integrates the Task Priority control framework first discussed in \cite{Simetti2016d}.

\subsection{Representation of Cooperation Models}
\label{sec:representation}

An AND/OR graph $G(N,H)$ is a data structure where $N$ is a set of $n_1, \ldots, n_{|N|}$ nodes and $H$ is a set of $h_1, \ldots, h_{|H|}$ hyper-arcs.
Nodes in $N$ define reachable \textit{states} in the graph, whereas hyper-arcs in $H$ define \textit{transition} relationships among those states.
Each hyper-arc $h_i \in H$ defines a \textit{many-to-one} transition relationship between a set of $|c|$ child nodes $c(h_i) = (n_{h_{i,1}}, \ldots, n_{h_{i,|c|}})$ and a parent node $p(h_i) = n_k$.
The relationships between the transitions of a hyper-arc are considered to be in logical \textit{and}, while the relationships between different hyper-arcs of the same parent node are in logical \textit{or}.
Both nodes and hyper-arcs can be associated with costs, namely $w_{n_1}, \ldots, w_{n_{|N|}}$ and $w_{h_1}, \ldots, w_{h_{|H|}}$.

In FlexHRC, each hyper-arc $h_i$ models a set $A_i$ of actions $a_1, \ldots, a_{|A_i|}$, and an action $a_j \in A_i$ can be assigned either to a human or a robot, according to their capabilities, during the cooperation process.
If the order in which to execute actions in $A_i$ is important, $A_i$ is defined as an \textit{ordered} set of actions such that ${A}_i = (a_1, \ldots, a_{|A_i|}; \preceq)$, meaning that a \textit{temporal} sequence is assumed in action execution in the form $a_1 \preceq a_2 \preceq \ldots \preceq a_{|A_i|}$.
Initially, all actions $a_j\in A_i$ are labeled as \textit{unended}, i.e., $\neg e(a_j)$.
When an action $a_j$ has been executed by the human or the robot, it is labeled as \textit{ended}, i.e., $e(a_j)$.
For all actions in $A_i$, if $e(a_j)$ holds then $h_i$ is \textit{done}, and the notation $d(h_i)$ is used.
If an ordering is induced, $d(h_i)$ holds if and only if also the temporal execution sequence is satisfied.
By design, all sets $A_i$ are different from each other.

A node can be either solved or unsolved.
A node $n_k \in N$ is \textit{solved}, specified with $s(n_k)$, if there is at least one hyper-arc $h_i \in H$ such that $p(h_i) = n_k$, $d(h_i)$ holds, and for all $n_l \in c(h_i)$ it holds that $s(n_l)$.
A node $n_k$ is \textit{unsolved} otherwise, and specified with $\neg s(n_k)$.
\textit{Leaves} in $G$ are initialized as solved or unsolved, depending on the initial state of the cooperation.

An AND/OR graph $G$ is traversed from leaves to the \textit{root} node $n_r \in N$.
When $s(n_r)$ holds, then $G$ is \textit{solved}, i.e., $s(G)$.
During the traversal procedure, a node $n_k \in N$ is \textit{feasible}, i.e., $f(n_k)$ holds if there is at least one hyper-arc $h_i \in H$ such that $p(h_i) = n_k$ and for all nodes $n_l \in c(h_i)$ it holds that $s(n_l)$. In this case $h_i$ is labeled as \textit{active}, i.e., $a(h_i)$. Otherwise, $n_k$ is \textit{unfeasible}, i.e., $\neg f(n_k)$.
While the cooperation unfolds, there is a set of active hyper-arcs $H_a \subset H$ in $G$.

Let us define the \textit{graph representation state} $S_G$ as the set of all feasible nodes and active hyper-arcs in $G$.
$S_G$ defines possible action alternatives for the human or the robot during the cooperation process.
A cooperation model $M$ is a ordered sequence of $|M|$ actions, such that $M = (a_1, \ldots, a_{|M|}; \preceq) \subset S_G$, corresponding to an allowed cooperation path $P$ in $G$.
Each cooperation path is associated with a \textit{traversal} cost, namely $cost(P)$, which defines how effortful executing $P$ is, on the basis of the involved node and hyper-arc weights.   
At any given time instant, there is one current cooperation model $M_c$ and one current cooperation path $P_c$.






\begin{algorithm}[!t]
\begin{algorithmic}[1]
\caption{Setup()}
\label{alg:setup}
\Require {A description of an AND/OR graph $G=(N,H)$}
\Ensure {A data structure encoding $G$}
\State $G \leftarrow$ loadDescription()
\State $s(G) \leftarrow$ \textit{false}
\ForAll{$n \in N$}
\State updateFeasibility($G$, $n$)
\EndFor
\State $\mathcal{P} \leftarrow$ generateAllPaths($G$)
\State $n^* \gets$ findSuggestion($\mathcal{P}$)
\end{algorithmic}
\end{algorithm}

\begin{algorithm}[!t]
\begin{algorithmic}[1]
\caption{updateFeasibility()}
\label{alg:update_feasibility}
\Require {An AND/OR graph $G=(N,H)$, a node $n \in N$}
\Ensure {$f(n)$ or $\neg f(n)$}
\If{$f(n)$ = \textit{true}}
\State \textit{return} 
\EndIf
\If{$c(n) = \emptyset$}
\State $f(n) \leftarrow$ \textit{true}
\State \textit{return}
\EndIf
\ForAll{$h \in H$ such that $p(h) = n$}
\State allChildNodesSolved $\leftarrow$ \textit{true}
\ForAll{$m \in c(h)$}
\If{$s(m)$ = \textit{false}}
\State allChildNodesSolved $\leftarrow$ \textit{false}
\EndIf
\EndFor
\If{allChildNodesSolved = \textit{true}}
\State $a(h)$ $\leftarrow$ \textit{true}
\State $f(n) \leftarrow$ \textit{true}
\State \textit{return}
\EndIf
\EndFor
\State $f(n) \leftarrow$ \textit{false}
\State \textit{return}
\end{algorithmic}
\end{algorithm}

\begin{algorithm}[!t]
\begin{algorithmic}[1]
\caption{generateAllPaths()}
\label{alg:generate_allpaths}
\Require {An AND/OR graph $G=(N,H)$}
\Ensure {The set $\mathcal{P}$ of all cooperation paths}
\State $\mathcal{P} \leftarrow \emptyset$
\State $P \leftarrow$ initNewPath()
\State $\mathcal{P} \leftarrow \mathcal{P} \cup P$
\ForAll{$n \in N$}
\State $e(n) \leftarrow$ \textit{false}
\EndFor
\State addNode($P$, $n_r$)
\While{\textit{true}}
\If{$\forall P \in \mathcal{P}$ it holds that $\forall n \in P$, $e(n)$ = \textit{true}}
\State \textit{return} $\mathcal{P}$
\Else
\State $P, n \leftarrow$ getUnexploredNode($\mathcal{P}$)
\State generatePath($G$, $n$, $P$)
\EndIf
\EndWhile
\end{algorithmic}
\end{algorithm}

\begin{algorithm}[!t]
\begin{algorithmic}[1]
\caption{generatePath()}
\label{alg:generate_path}
\Require {An AND/OR graph $G=(N,H)$, the current node $n$, a cooperation path $P$}
\Ensure {A valid cooperation path $P$}
\State updatePathCost($P$, $n$)
\State $e(n) \leftarrow$ \textit{true}
\ForAll{$h \in H$ such that $p(h) = n$}
\If{$|h| = 0$}
\State \textit{return}
\ElsIf{$|h| = 1$}
\State addArc($P$, $h$)
\ForAll{$m \in c(h)$}
\State addNode($P$, $m$)
\EndFor
\State \textit{return}
\ElsIf{$|h| > 1$}
\State $P' \leftarrow P$
\ForAll{$h \in H$ such that $p(h) = n$}
\State $P \leftarrow$ initNewPath($P'$)
\State $\mathcal{P} \leftarrow \mathcal{P} \cup P$
\State addArc($P$, $h$)
\ForAll{$m \in c(h)$}
\State addNode($P$, $m$)
\EndFor
\EndFor
\State \textit{return}
\EndIf
\EndFor
\end{algorithmic}
\end{algorithm}
All available cooperation models and cooperation paths are determined \textit{offline} for \textit{online} use, and maintained by the \textit{Task Representation} module.   
Algorithm \ref{alg:setup} starts the process. 
A description of $G$ is loaded from file (line 1), and $G$ is marked as unsolved (line 2).
Then, all node feasibility states are determined (line 4), all possible cooperation paths are generated (line 6) and the first node to solve $n^*$ is determined (line 7).

Feasibility check is performed on each node in $N$. 
The process is described in Algorithm \ref{alg:update_feasibility}.
If a node $n \in N$ is feasible, then it remains feasible (line 2).
If a node does not have any hyper-arcs connecting it to child nodes, it does not have preconditions, and therefore it is feasible (line 5).
Otherwise, the algorithm looks for active hyper-arcs (lines 8 to 20): if there is at least one hyper-arc $h$ for which all child nodes are solved, then the hyper-arc is active (line 16) and $n$ is feasible (line 17), otherwise, the hyper-arc is ignored (lines 9 to 14). If the node has no active hyper-arc, then it is not feasible (line 21).

When Algorithm \ref{alg:update_feasibility} is complete, the graph representation state $S_G$ is available, and it is possible to determine all available cooperation paths.
This is done by Algorithm \ref{alg:generate_allpaths}, which is a variation of a depth-first traversal procedure for AND/OR graphs.
The set of cooperation paths $\mathcal{P}$ and an empty path $P$ are defined (lines 1 to 3).
All nodes are marked as unexplored (line 5) and the root node $n_r$ is added to path $P$ (line 7).
Then, the procedure iterates calling Algorithm \ref{alg:generate_path} on the unexplored nodes (lines 12 and 13) until all nodes are explored (lines 9 and 10).

Algorithm \ref{alg:generate_path} proceeds along a single cooperation path $P$, starting from the current node. The cost of $P$ is updated and the node is marked as explored (lines 1 and 2).
Then, all its child nodes are determined.
If the node does not have child nodes, the exploration of the path $P$ from node $n$ is completed (lines 4 and 5).
Otherwise, if all the child nodes of $n$ belong to the same hyper-arc $h$, the hyper-arc is added to $P$ (line 7) and the child nodes are added to $P$ for later exploration (line 9). Finally, if the child nodes of $n$ belong to more than one hyper-arc, a new path is created for each hyper-arc, as a copy of the current path (lines 15 and 16). The different hyper-arcs and the corresponding child nodes are added to the new paths (lines 17 and 19) for later exploration.
When these procedures end, FlexHRC is ready for online cooperation.
Depending on the optimal cooperation path $P^*$, i.e., the one minimizing the overall cost depending on node and hyper-arc weights, the robot may start moving or waiting for human actions.

\subsection{Classification of Human Actions}
\label{sec:determining}

Determining human actions is a task carried out jointly by the \textit{Human Action Recognition} module and the \textit{Planner} module (Figure \ref{fig:SysArchitecture}).
Whilst the former performs \textit{gesture recognition} using inertial data collected at the human co-worker's wrist, the latter integrates this information to determine how it fits with actions in hyper-arcs.

The \textit{Human Action Recognition} module employs a system for gesture recognition and classification first described in \cite{Bruno2013,Bruno2014}.
The approach identifies two phases, namely an offline training phase, when a set $\mathcal{G}$ of $g_1, \ldots, g_{|\mathcal{G}|}$ gesture models are created from a purposely obtained training set of inertial data, and an online phase, when human motions are classified on the basis of the gesture models in $\mathcal{G}$.

As described in \cite{Bruno2013}, after a data filtering process aimed at isolating \textit{gravity} and \textit{body acceleration} as features, the modeling process relies on Gaussian Mixture Modeling (GMM) and Gaussian Mixture Regression (GMR) to compute an \textit{expected} regression curve and the covariance matrix for each $g \in \mathcal{G}$.
Once the regression curve for a gesture model $g$ is obtained, the number of data points in it needs not to be the same as that of trials in the training set.
This is of the utmost importance for coping with computational requirements in the online phase.
This modeling procedure requires in input the number of Gaussian functions to use, which is feature-dependent.
In order to determine it, a hybrid procedure based on the k-means algorithm and the silhouette clustering  metric is adopted \cite{Rousseeuw1987}. 

While the human-robot cooperation process unfolds, \textit{Human Activity Recognition} executes a number of steps, in part similar to the modeling procedure in the offline phase.
Once the online inertial data stream has been processed to extract gravity and body acceleration features (typically focusing on a time window depending on gesture model lengths), \textit{gesture recognition} is performed by comparing those features against the models in $\mathcal{G}$, thereby labeling data with a gesture symbol.
To this aim, a specific distance metrics is adopted, e.g., the well-known Mahalanobis distance and the maximization of the so-called \textit{possibilities}, to take into account the variability associated with gesture models. 

\begin{figure}[!t]
\centering
\includegraphics[width=\columnwidth]{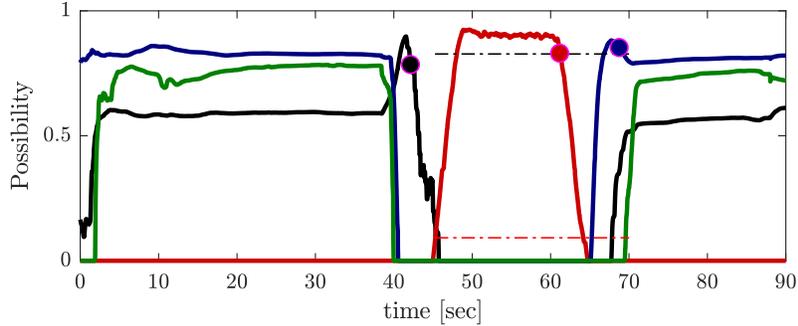}
\caption{An example of action possibility evolutions for a human-robot cooperation task.}
\label{fig:HAR_Possibilty}
\end{figure}
In this work, possibilities are exploited to determine which gesture has been executed and when they can be considered as \textit{ended} \cite{Bruno2014}.
A possibility measure quantifies the correlation between the online data stream and each gesture model. 
Let us assume that, at a given time instant, the considered time window contains an inertial data pattern related to a gesture model $g$.
Intuitively, the correlation between the time window and the correct gesture model reaches its peak when the former is in perfect overlap with the model.
When a full overlap is still to be reached, the possibility value increases towards the peak value, whereas, if full overlap is passed, the possibility value decreases.
An example is given in Figure \ref{fig:HAR_Possibilty}, where possibility values for four human gestures are shown, using different colors (see Section \ref{sec:evaluation} for a more detailed description). 
Focusing on the \textit{red} pattern, it can be seen that the associated possibility value is zero for the first $45$~s, then it jumps to reach almost $1$, and afterwards it slightly decreases.
There might be small oscillations in possibility values, which might cause local maxima.
To overcome this issue, a threshold is introduced to find the (semi-global) maximum of the possibility pattern.
If this threshold is too low, the delay introduced by gesture recognition will be high.
When, after the peak, the possibility reaches a value that amounts to the threshold (currently set at $90\%$ of the peak possibility value, red dot in the Figure at $62$~s), the corresponding gesture $g$ is considered as executed, subject to the fact that it is the highest among all other model possibilities.

Once a gesture has been detected, the corresponding label is given to the \textit{Planner} module, which interacts with \textit{Task Representation} to update the graph state $S_G$.
Gestures are used to confirm the execution of human actions as provided for in hyper-arcs.

When a human action $a_j$ (corresponding to a gesture $g$) is detected by the \textit{Human Activity Recognition} module, \textit{Task Representation} determines whether an $A_i$ exists such that $a_j \in A_i$, and $A_i$ corresponds to an active hyper-arc.
If this is the case, and $A_i$ is associated with an $h_i$ part of $P_c$, the cooperation continues with $M_c$, and the next action is suggested to minimize costs along $P_c$; if $A_i$ corresponds to an hyper-arc not part of $P_c$, the user switched to another cooperation model, and the corresponding cooperation path is set as new $P_c$.
Lastly, if $a_j$ is part of multiple active hyper-arcs, FlexHRC enters an \textit{ambiguous} mode and waits for further inputs (i.e., other actions) to determine which cooperation path $P$ in $G$ is involved.
Considering a set $A_i$ corresponding to an active hyper-arc $h_i$, when there are no ordering constraints on $A_i$, the order of action execution therein (and therefore, the order in which human gestures are detected) is not relevant for $h_i$ to be labeled as done.
Conversely, when an ordering is imposed, FlexHRC assumes that the active hyper-arc is done when the various actions are executed in strict sequence.
Given an hyper-arc $h_i$, to understand which actions of $A_i$ have been executed, predicates $e(a_j)$ are set to true appropriately.
When a gesture $g$ is recognized, the \textit{Planner} compares it with the first \textit{unended} action in all active hyper-arcs.
If no corresponding action is found in a hyper-arc, the latter is marked as \emph{inactive}.   

It is noteworthy that a similar updating mechanism occurs for actions executed by the robot, as discussed in Section \ref{sec:taskpriority}.
As a result of this update, a planning process can occur to determine the next suggested action for humans and robots.

\subsection{Planning the Next Human or Robot Action}
\label{sec:planning}

Whenever the graph state $S_G$ is updated as a consequence of human or robot actions, the next node to solve $n^*$ in the current cooperation path $P_c$ (or in a new one, in case of a human switching to a new one) can be defined.

\begin{algorithm}[!t]
\begin{algorithmic}[1]
\caption{NextSuggestedNode()}
\label{alg:run}
\Require {An AND/OR graph $G$, the last solved node $n \in N$}
\Ensure {An updated AND/OR graph $G$, the next node to solve $n^*$}
\State $loop \leftarrow$ \textit{true}
\State $s(n) \leftarrow$ \textit{true}
\While{$loop$ = \textit{true}}
\If{$n$ = $n_r$}
\State $loop \leftarrow$ \textit{false}
\EndIf
\ForAll{$m \in N$}
\State updateFeasibility($m$)
\EndFor
\State updateAllPaths($\mathcal{P}$)
\State $n^* \leftarrow$ findSuggestion($\mathcal{P}$)
\State \textit{return}
\EndWhile
\State $s(G) \leftarrow$ \textit{true}
\State \textit{return}
\end{algorithmic}
\end{algorithm}

\begin{algorithm}[!t]
\begin{algorithmic}[1]
\caption{updateAllPaths()}
\label{alg:update_paths}
\Require {The set $\mathcal{P}$ of all cooperation paths, the last solved node $n \in N$}
\Ensure {An updated set $\mathcal{P}$}
\State $P^u \leftarrow \emptyset$ 
\State $P^u \leftarrow$ findPathsToUpdate($n$)
\ForAll{$P \in P^u$}
\State $cost(P) \leftarrow$ updateCost()
\EndFor
\end{algorithmic}
\end{algorithm}
This is done by Algorithm \ref{alg:run}.
The Algorithm loops indefinitely until the root node $n_r$ is reached, and therefore $s(G)$ holds (lines 4, 5 and 14). 
In the meantime, it updates all feasibility states (lines 7-9) as well as cooperation paths (line 10), and provides a suggested next node $n^*$ to solve (line 11).

Whilst feasibility updates are managed by Algorithm \ref{alg:update_feasibility}, cooperation path updates are done as described in Algorithm \ref{alg:update_paths}. 
Here, the set $P^u$ of all paths to update is determined (line 2) as those containing the last solved node $n$.
Then, for each path $P$, its associated cost is updated as follows:
\begin{equation}
cost(P) = cost(P) - (w_n + h_n^m - w_h),
\end{equation}
where $w_n$ is the weight associated with $n$, $h_n^m$ is the maximum weight of the hyper-arcs connecting any parent node to $n$, while $w_h$ is the weight of the hyper-arc connecting any parent node to $n$ in $P$.

\begin{algorithm}[!t]
\begin{algorithmic}[1]
\caption{findSuggestion()}
\label{alg:find_suggestion}
\Require {The set $\mathcal{P}$ of all cooperation paths }
\Ensure {The next node to solve $n^*$}
\State $P^* \leftarrow$ findOptimalPath($\mathcal{P}$)
\ForAll{$n \in P^*$}
\State $n^* \leftarrow$ findOptimalNode()
\EndFor
\State \textit{return}
\end{algorithmic}
\end{algorithm}
Finally, suggestions for the next node $n^*$ are determined by the procedure in Algorithm \ref{alg:find_suggestion}.
First, the optimal cooperation path $P^*$ is determined, such that it is characterized by the minimum cost (line 1).
Then, for all nodes in $P^*$, the first node $n$ is found such that $f(n)$ and $\neg s(n)$ hold, which is labeled as $n^*$.  

On the basis of $n^*$, the actions to expect from the human  or to be executed by the robot are determined, and the procedure continues.

\subsection{Robot Control and Action Execution}
\label{sec:taskpriority}

This Section describes in detail the Task Priority framework integrated within the \textit{Controller} module of Figure \ref{fig:SysArchitecture}, and in particular its flexibility in decoupling control from action planning as performed by the \textit{Planner} module is discussed.

In the following paragraphs, the robot configuration vector is named $\vec{c} \in \mathbb{R}^n$ and contains the robot degrees of freedom (DOF), e.g., joint positions, while the robot velocity vector is named $\dot{\vec{y}} \in \mathbb{R}^n$, and represents the controls to actuate the robot, e.g. joint velocities.

In Task Priority schemes, it is necessary first to define control objectives and the boundaries associated with reference and control objective values.
A control objective $o$ expresses what the robot needs to achieve: for example, reaching a desired position for the end-effectors, maintaining an object in the camera's field of view, or avoiding arm joint limits.
Let us consider a scalar variable $x(\vec{c})$ depending on the current robot configuration $\vec{c}$.
Let us call a \emph{scalar equality control objective} $o_e$ a requirement such that, for $t \to \infty$, $x(\vec{c}) = x_0$, where $x_0$ is a given reference value.
Also, a \emph{scalar inequality control objective} $o_i$ is a requirement such that, for $t \to \infty$, $x(\vec{c}) > x_{min}$ or $x(\vec{c}) < x_{max}$, where $x_{min}$ and $x_{max}$ serve as lower and upper thresholds for the values scalar variables can assume.

For each control objective $o$, a \emph{feedback reference rate} $\dot{\bar{x}}_o$ is defined, whose role is to drive the related variable $x(\vec{c})$ toward an arbitrary point $x^*$ inside the admissible state space where $o$ is satisfied.
The mapping between the task velocity scalar $\dot{x}_o$ and the system velocity vector $\dot{\vec{y}}$ is given by the Jacobian relationship:
\begin{equation*}
\dot{x}_o = \vec{J}_o^T \dot{\vec{y}}.
\end{equation*}
The \emph{control task} $\tau_o$, associated to the objective $o$, is defined as the need of minimizing the difference between the actual task velocity $\dot{x}_o$ and the feedback reference rate $\dot{\bar{x}}_o$.

Control objectives may or may not be relevant in a given situation.
In FlexHRC, this information is managed using a set of activation functions.
Activation functions are aimed at dynamically tuning the system's behavior at runtime, for instance on the basis of sensory data.
For example, considering the problem of avoiding an obstacle with one of the robot links, then the related task would be relevant only when the link is close to the obstacle.
This task should not over-constrain the robot whenever the link is sufficiently far away from the obstacle.
This example is particularly relevant in the context of this work, and will be examined in the next Section.
Motivated by the above considerations, let us define a prototype for activation functions such that:
\begin{equation}
\label{eq:actfun}
\alpha(x) = \alpha_o(x),
\end{equation}
where $\alpha_o(x) \in [0, 1]$ is a continuous sigmoid function of a scalar objective variable $x$, whose value is zero within the validity region of the associated control objective $o$.
 
After control objectives, reference values, lower and upper thresholds and the corresponding activation functions are identified, their respective priority must be determined.
The approach of Task Priority schemes is to define $p$ priority levels so that:
(i) each task is assigned to one priority level;
(ii) low priority tasks are inhibited from interfering with high priority ones;
(iii) different scalar objectives assigned to the same priority level can be grouped in a (possibly multidimensional) control objective. 
Assuming a priority level $k$ with $m$ scalar control tasks (and therefore $m$ control objectives), the following vectors and matrices are defined.
\begin{itemize}
\item $\dot{\bar{\vec{x}}}_k \triangleq \begin{bmatrix}\dot{\bar{x}}_{1,k}, \ldots, \dot{\bar{x}}_{m,k}\end{bmatrix}$ is the stacked vector of all the reference rates.
\item $\mat{J}_k$ is the Jacobian relationship  expressing the current rate of change of the $k$-th task vector $\begin{bmatrix}\dot{{x}}_{1,k}, \ldots, \dot{{x}}_{m,k}\end{bmatrix}$ with respect to the system velocity vector $\dot{\vec{y}}$;
\item $\mat{A}_k \triangleq \textrm{diag}(\alpha_{1,k}, \ldots, \alpha_{m,k})$ is the diagonal matrix of all the activation functions in the form of \eqref{eq:actfun}.
\end{itemize}
With these definitions, the control problem is to find the system's velocity reference vector $\dot{\bar{\vec{y}}}$ complying with the aforementioned priority requirements.
In order to compute such a vector, a Task Priority Inverse Kinematic (TPIK) procedure has been proposed in \cite{Simetti2016d}.
Here, it would suffice to describe the single regularization and optimization step, which unfolds iteratively taking into account all lower priority tasks.   
The manifold of solutions at the $k$ level is:
\begin{equation}
\label{eq_rminproblem}
S_k \triangleq \left\{ \arg \mathrm{R\textrm{-}}\min_{\dot{\bar{\vec{y}}} \in S_{k-1}} \left\| \mat{A}_k (\dot{\bar{\vec{x}}}_k - \mat{J}_k \dot{\bar{\vec{y}}}) \right\|^2 \right\},
\end{equation}
where $S_{k-1}$ is the manifold of solutions of all the previous tasks in the hierarchy, with $S_0 \triangleq \mathbb{R}^n$.
Since $k$ is increased at each step, the recursion stops when $k = p$.
The notation $\mathrm{R\textrm{-}}\min$ is used to highlight the fact that the minimization must be regularized to avoid algorithm singularities during transitions between pairwise tasks.
This regularization mechanism and the resulting TPIK algorithm are duly reported in \cite{Simetti2016d} and will be omitted here for the sake of brevity.

From the robot control standpoint, an action $a$ in input to the \textit{Controller} module in Figure \ref{fig:SysArchitecture} can be defined as a prioritized list of $m$ control objectives $o_1, \ldots, o_m$ and the associated control tasks $\tau_1, \ldots, \tau_m$, to be managed \emph{concurrently}.
As an example, a manipulation action could include the following list of objectives, in order of priority:
\begin{enumerate}
\item[$o_1$] Arm joint limits.
\item[$o_2$] Arm obstacle avoidance.
\item[$o_3$] Arm manipulability.
\item[$o_4$] End-effector linear position control.
\item[$o_5$] End-effector angular position control.
\item[$o_6$] Arm preferred pose.
\end{enumerate}

Thanks to the proposed TPIK scheme, safety-oriented objectives such as \textit{arm joint limits} and \textit{arm obstacle avoidance} can be given high priority in the hierarchy, and they can be deactivated whenever irrelevant, through the use of properly defined activation functions.

Two remarks can be made.
The first is that, in principle, an action $a$ embeds an arbitrary number of $m$ prioritized objectives $o_1, \ldots, o_m$, organized in different hierarchies.
Typically, the main difference between any two actions is the set of objectives needed to achieve their goals and possibly other prerequisite objectives, whereas safety-oriented tasks are common to all actions.
The second is that since actions are sequenced according to a cooperation model represented by an AND/OR graph, each action in hyper-arcs is defined with a few control objectives relevant to the action goal only.
	
Given the second remark, it is necessary to describe how transitions between two subsequent actions are implemented to achieve a safe yet natural robot behavior.
As discussed above, it is realistic to assume that actions are characterized by a common set of safety-oriented objectives, and differ only by a few action-specific objectives.
For the sake of argument, let us imagine a unified list made up of all control objectives of two actions $a_1$ and $a_2$, e.g., $o_1, \ldots, o_{m_1+m_2}$.
It is easy to imagine how, by a simple removal of some of the control objectives, the two initial sets can be easily determined.
To do so, activation functions in the form of \eqref{eq:actfun} are modified as:
\begin{equation}
\label{eq:actfun2}
\alpha(x,\vec{p}) = \alpha_o(x) \alpha_p(\vec{p}),
\end{equation}
where $\alpha_p(\vec{p}) \in [0, 1]$ is a continuous sigmoid function of a vector of parameters $\vec{p}$ external to the control task itself.
In particular, $\alpha_p(\vec{p})$ can be conveniently parametrized by the two subsequent actions, as well as the time elapsed from the start time $T_{start}$ of the current action, to obtain the desired activation/deactivation smooth transition function between sets of objectives. 

Once all the actions in cooperation models are defined, such a unified list of objectives can be easily built.
Safety-oriented tasks are common to all actions, and they will be at the same (high) priority levels.
As a consequence, for such tasks it will result that $\alpha_p(\vec{p}) = 1$.
All other tasks will be instead managed by activation functions in the form $\alpha_p(\vec{p})$, to activate/deactivate action-specific and prerequisite objectives. 

When a robot action is successfully executed, the graph state $S_G$ in \textit{Task Representation} is updated and the \textit{Planner} can be invoked to define the next action.

\section{Experimental Evaluation}
\label{sec:evaluation}

\subsection{Scenario}

In order to provide a quantitative and qualitative assessment of FlexHRC, the experimental setup comprises a dual-arm Baxter manipulator to perform cooperative operations, and an LG G watch R (W110) smartmatch worn at the human right wrist to acquire inertial data. 

Smartwatch data are collected via an LG G3 smartphone, which is paired with the smartwatch using Bluetooth and communicates with a workstation using WiFi. The smartphone acquires inertial measurements from the smartwatch at approximately $40$ Hz via a Java application, and then forwards them to the workstation, which computes at best effort.
The Indigo ROS-based architecture runs on a $64$ bit i5 $2.3$ GHz workstation, equipped with $4$ Gb RAM, and Ubuntu 14.04.1 with kernel 3.19.0.

\begin{figure}[!t]
\centering
\includegraphics[width=0.9\columnwidth]{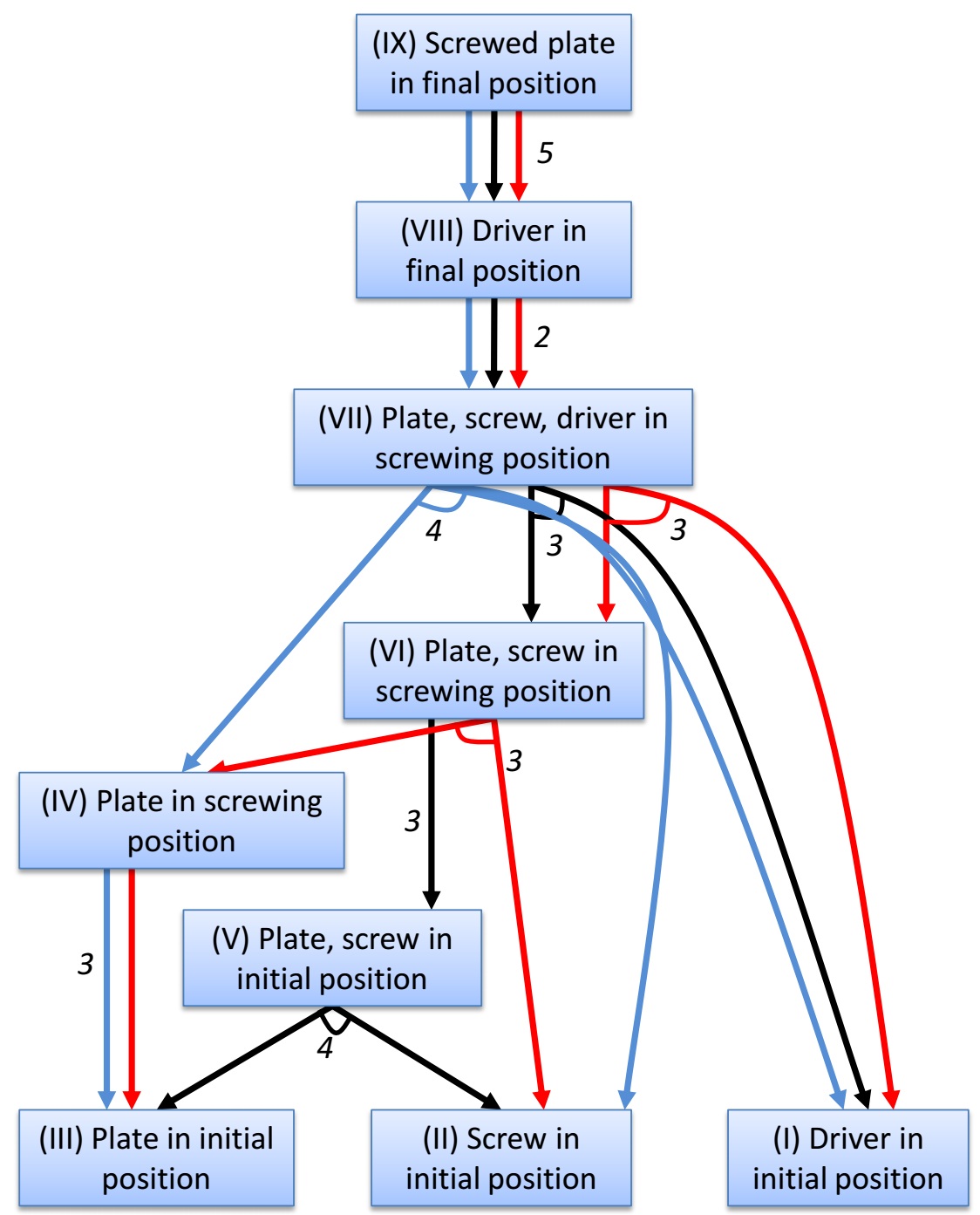}
\caption{The AND/OR graph representation of the screwing task: different colors (\textit{blue}, \textit{black} and \textit{red}) indicate different action sequences the cooperation can unfold in.}
\label{fig:AndOr}
\end{figure}
As a prototypical example, a screwing task has been considered.
A human wearing a smartwatch and Baxter face each other on the opposite sides of a table, where bolts, wooden plates and screwdrivers are located in \textit{a priori} known positions.
The goal of the task is to sink a bolt inside a wooden plate using a screwdriver, and place the assembled piece in a final goal position on the table.

Common sense and experience lead to the identification of three different cooperation models, referred to as $M_{blue}$, $M_{black}$ and $M_{red}$ in the following paragraphs, and represented in the corresponding AND/OR graph by three different paths, namely $\mathcal{P}_{blue}$, $\mathcal{P}_{black}$ and $\mathcal{P}_{red}$ (Figure \ref{fig:AndOr}), 
In this case, cooperation models are structured in advance. 
It is noteworthy that current work is devoted to learn cooperation paths from open-ended observations of how humans would behave if not instructed about how to cooperate with the robot, in order to extract the most natural interaction sequences from a human perspective.
However, this is out of the scope of the paper.
The resulting AND/OR graph has $|N| = 9$ states and $|H| = 9$ hyper-arcs.
The weight associated with each hyper-arc is specified by the estimated \textit{effort} needed by a human or a robot to complete the corresponding actions. 
As described above, such an effort does not necessarily consider only time, but may be a complex function taking into account also human preferences, ergonomic aspects of the operation, as well as its intrinsic difficulty.
However, such a function must be \textit{monotonic}, i.e., actions with associated low weights are to be preferred to actions characterized by high weights within each cooperation model. 
For this validation scenario, values have been \textit{a priori} determined through a test campaign with expert users.
As Figure \ref{fig:AndOr} shows, given this weights assignment, it follows that the optimal cooperation model is therefore $M_{blue}$, with a total expected cost of $cost(P_{blue}) = 14$.

\begin{figure*}[!t]
\centering
\begin{subfigure}[b]{0.2\textwidth}
\includegraphics[width=\textwidth]{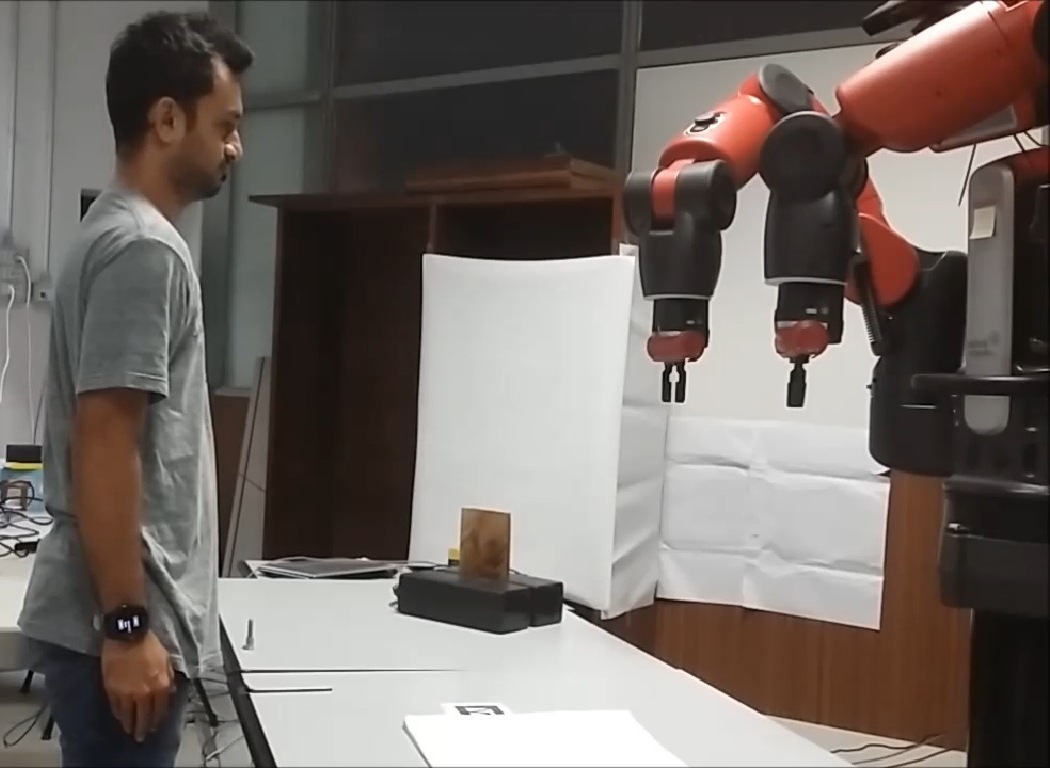}
\caption{}
\label{fig:p2_1}
\end{subfigure}
~
\begin{subfigure}[b]{0.2\textwidth}
\includegraphics[width=\textwidth]{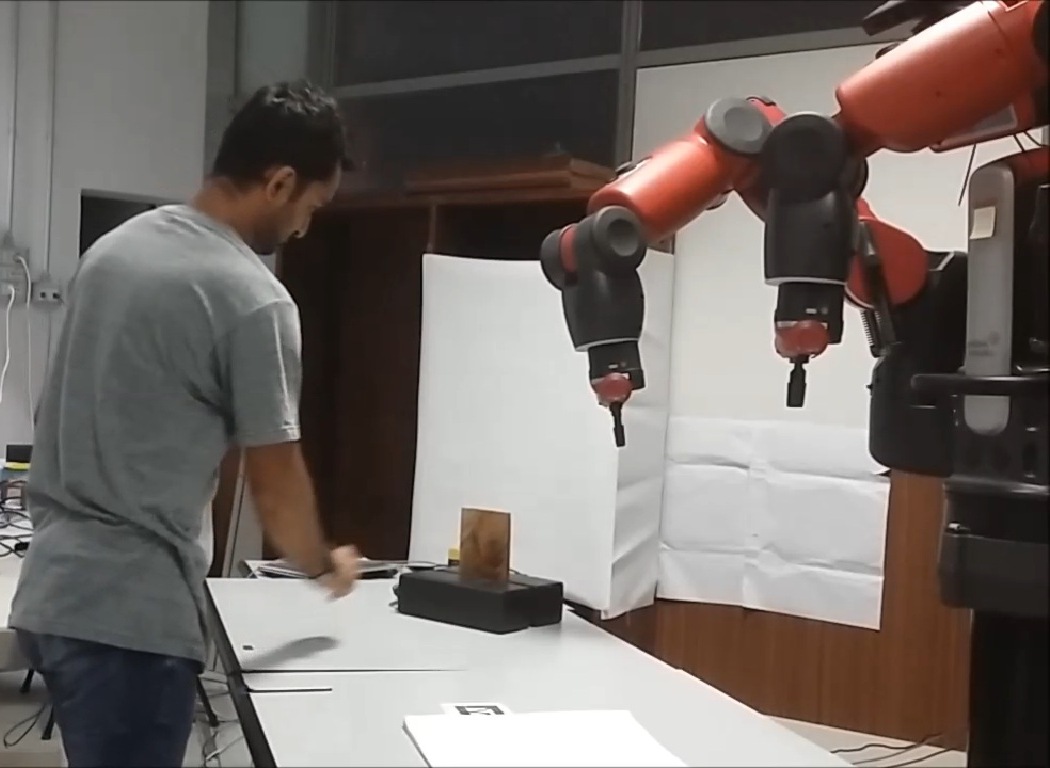}
\caption{}
\label{fig:p2_2}
\end{subfigure}
~
\begin{subfigure}[b]{0.2\textwidth}
\includegraphics[width=\textwidth]{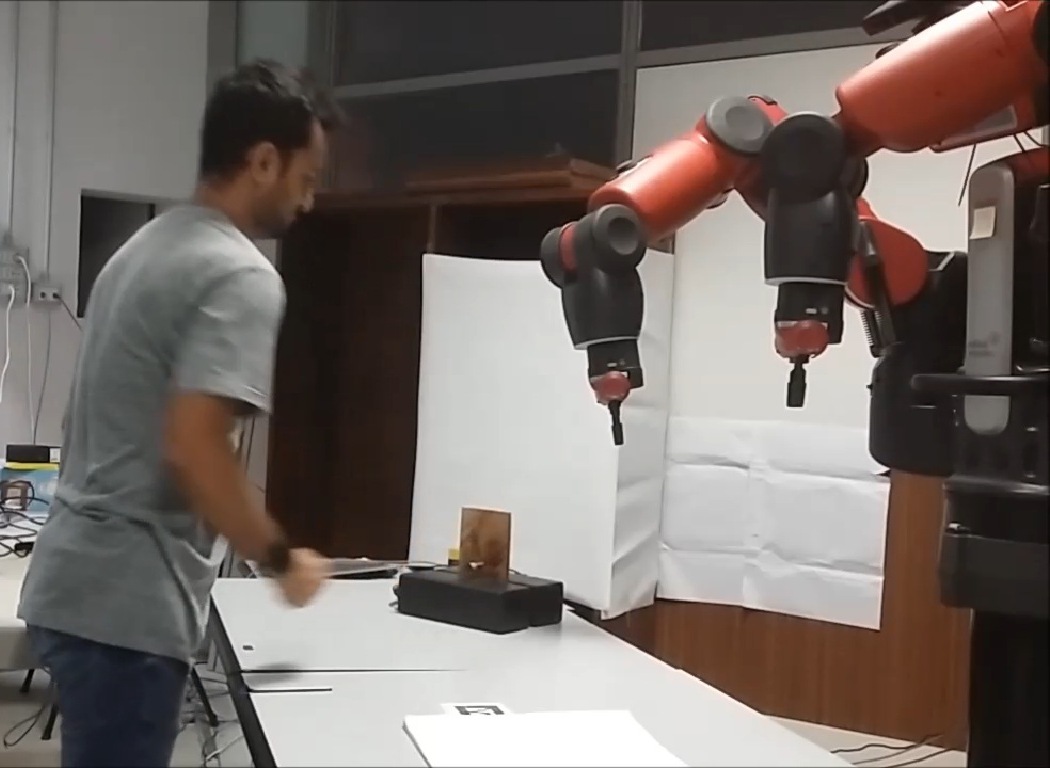}
\caption{}
\label{fig:p2_3}
\end{subfigure}
~
\begin{subfigure}[b]{0.2\textwidth}
\includegraphics[width=\textwidth]{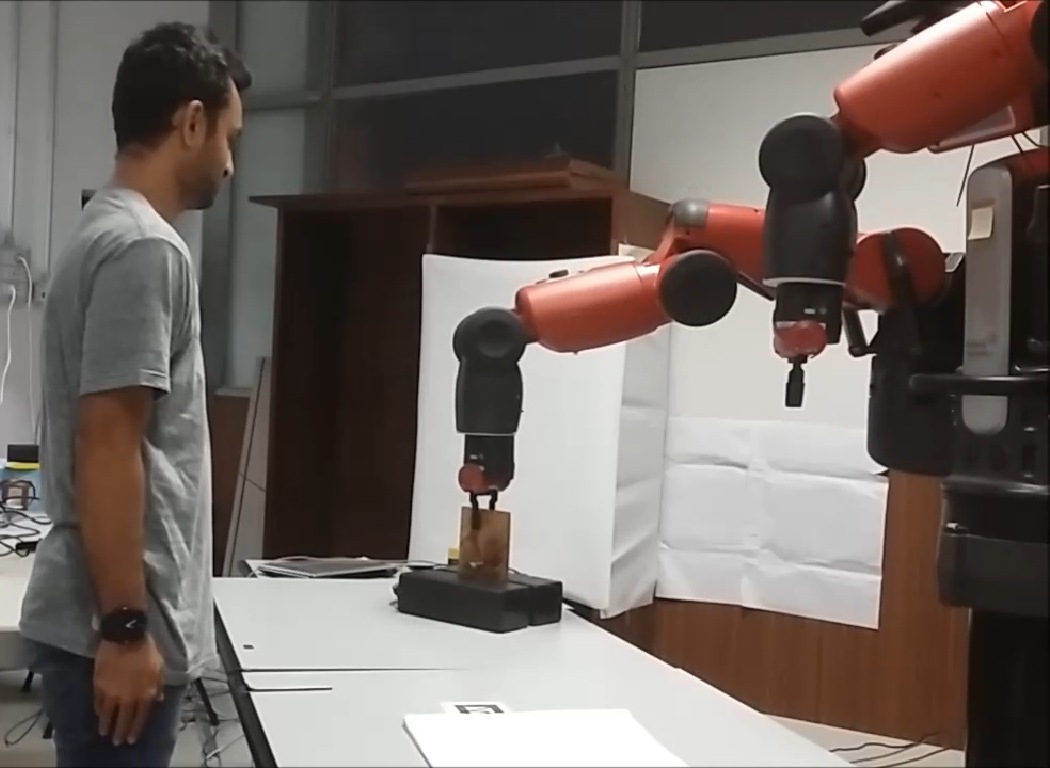}
\caption{}
\label{fig:p2_4}
\end{subfigure}

\begin{subfigure}[b]{0.2\textwidth}
\includegraphics[width=\textwidth]{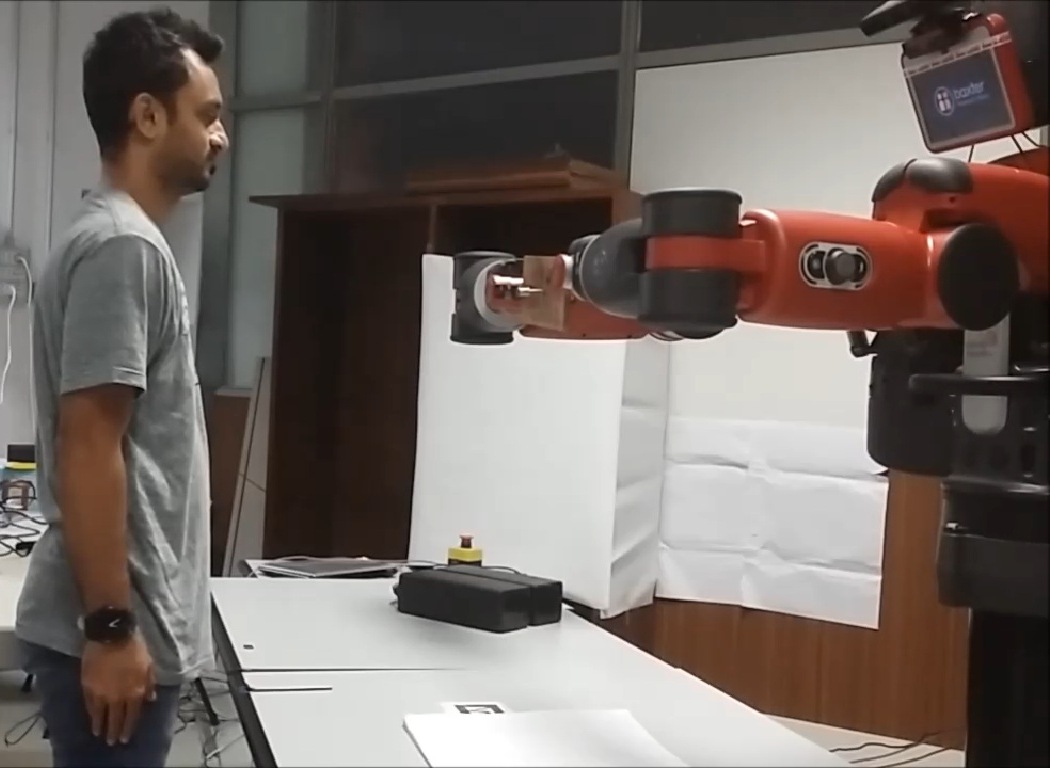}
\caption{}
\label{fig:p2_5}
\end{subfigure}
~
\begin{subfigure}[b]{0.2\textwidth}
\includegraphics[width=\textwidth]{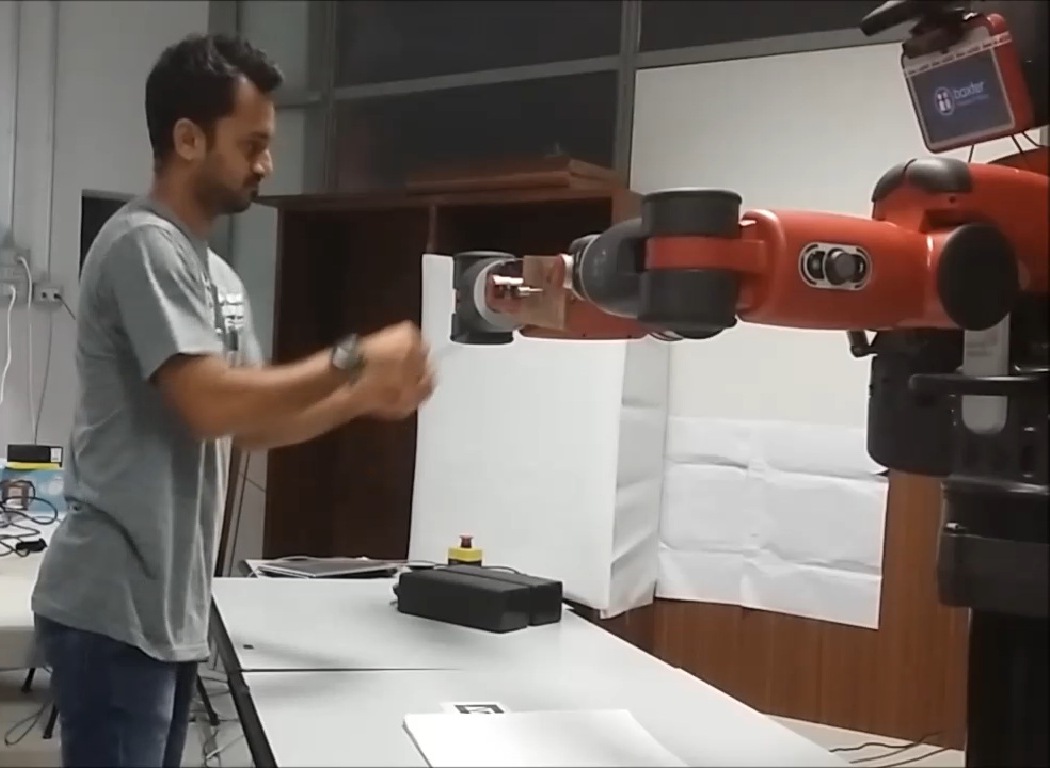}
\caption{}
\label{fig:p2_6}
\end{subfigure}
~
\begin{subfigure}[b]{0.2\textwidth}
\includegraphics[width=\textwidth]{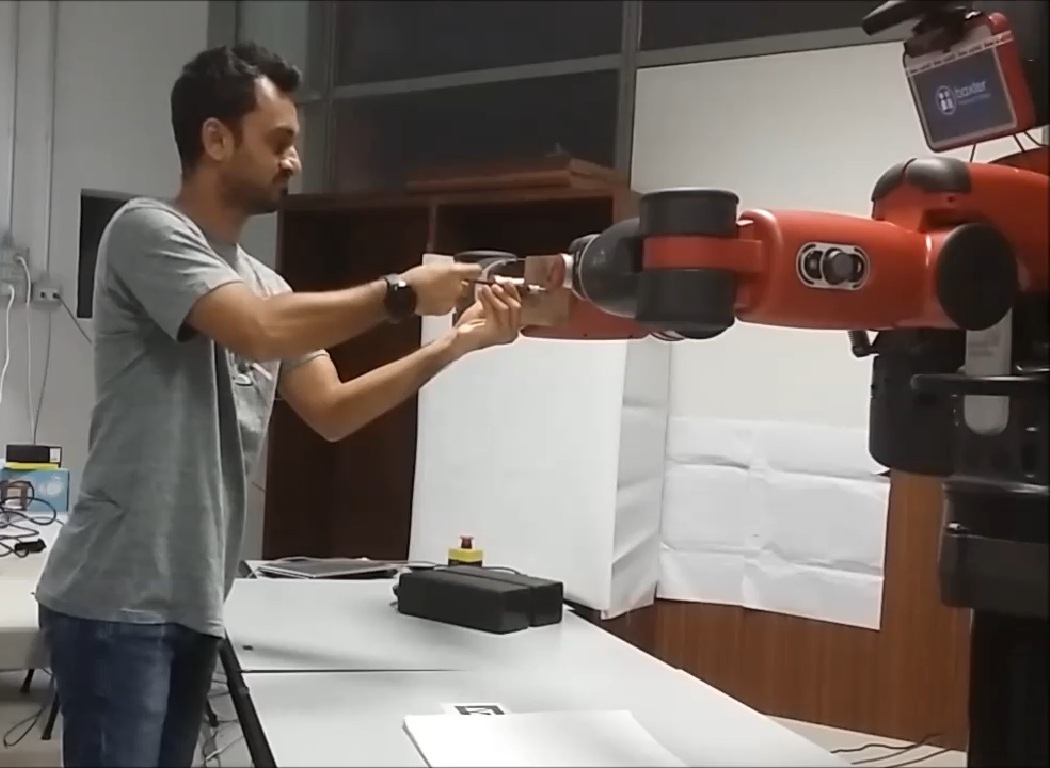}
\caption{}
\label{fig:p2_7}
\end{subfigure}
~
\begin{subfigure}[b]{0.2\textwidth}
\includegraphics[width=\textwidth]{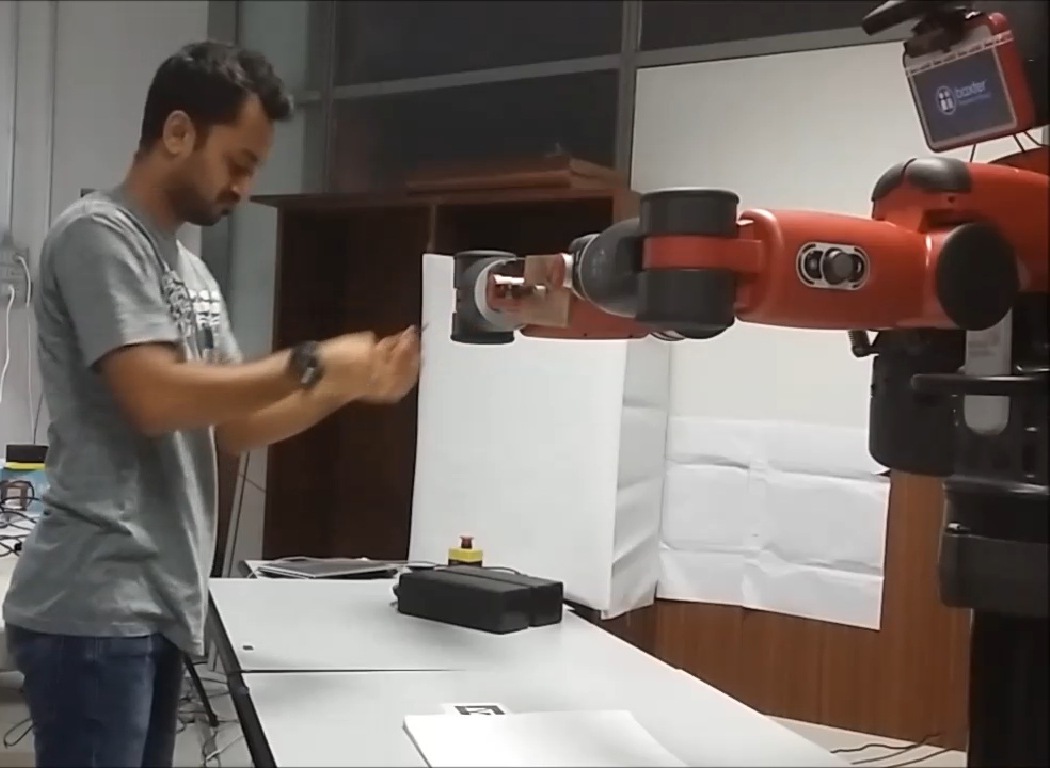}
\caption{}
\label{fig:p2_8}
\end{subfigure}

\begin{subfigure}[b]{0.2\textwidth}
\includegraphics[width=\textwidth]{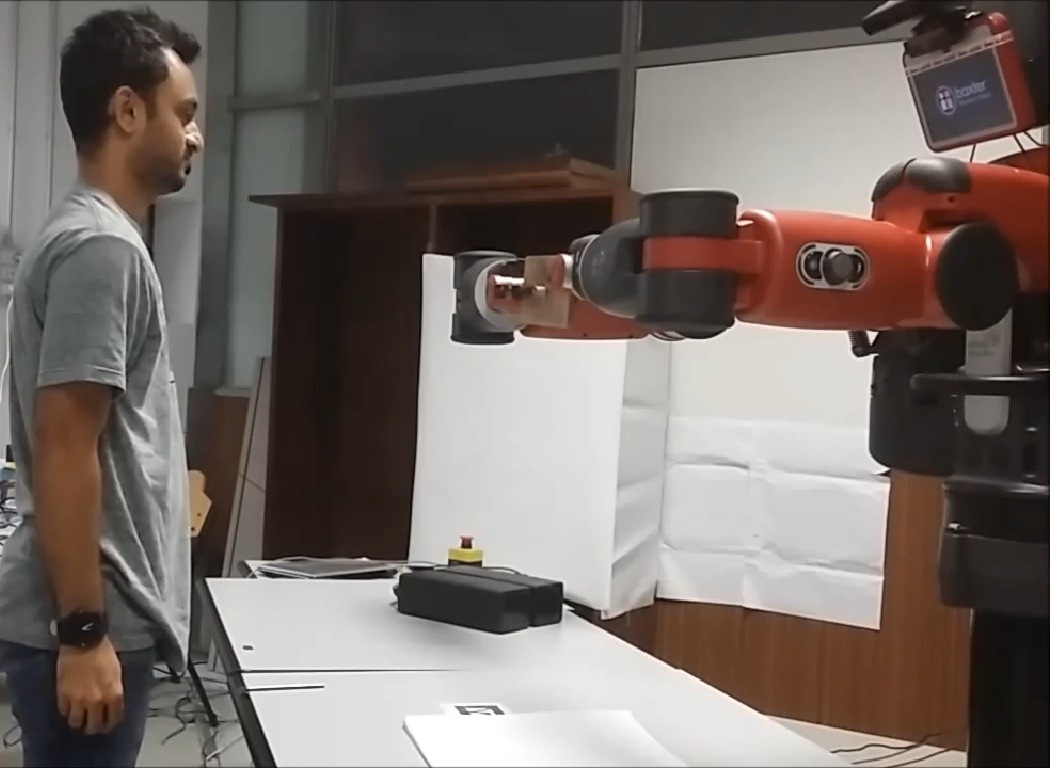}
\caption{}
\label{fig:p2_9}
\end{subfigure}
~
\begin{subfigure}[b]{0.2\textwidth}
\includegraphics[width=\textwidth]{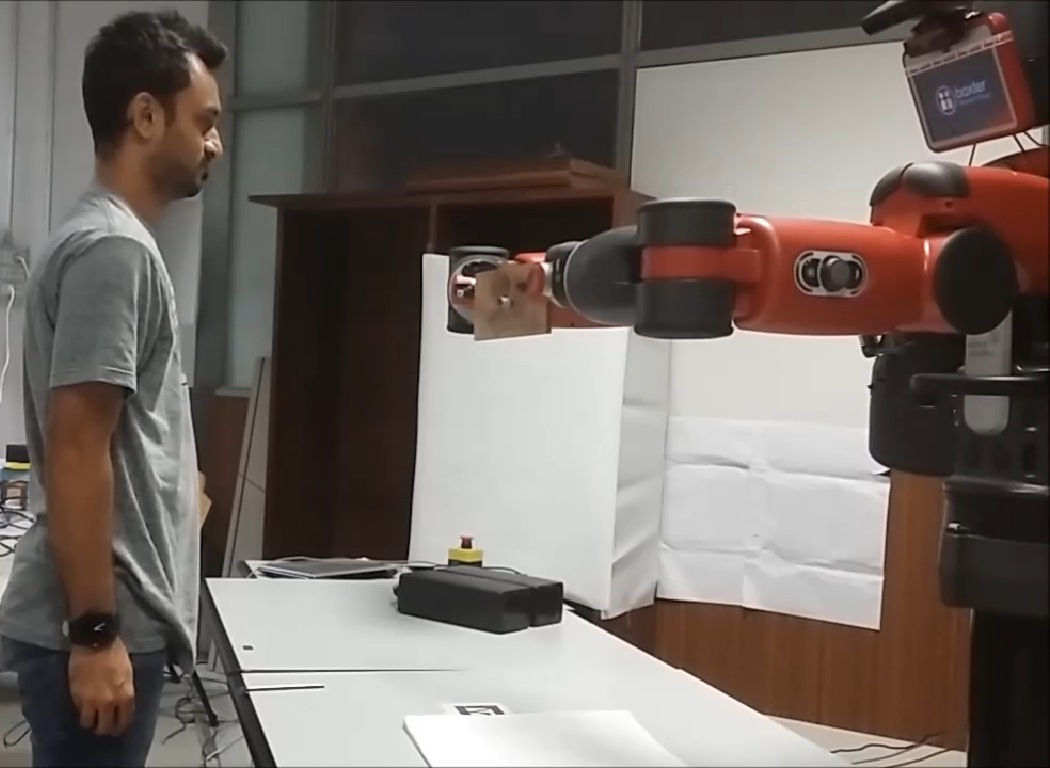}
\caption{}
\label{fig:p2_10}
\end{subfigure}
~
\begin{subfigure}[b]{0.2\textwidth}
\includegraphics[width=\textwidth]{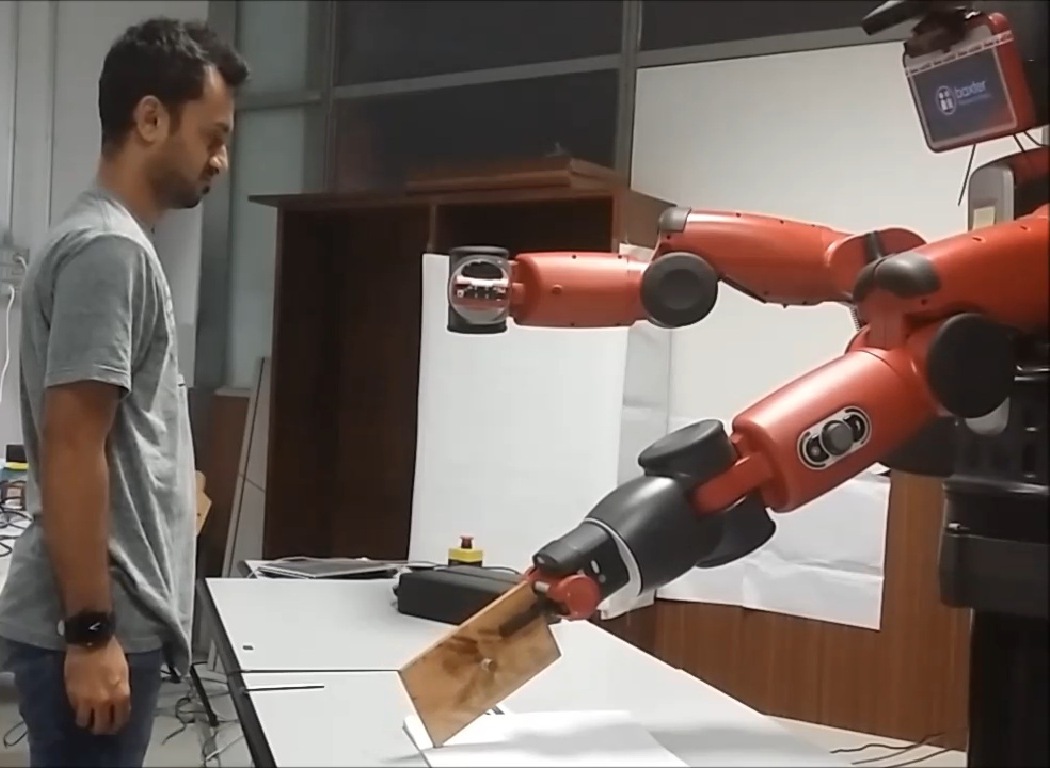}
\caption{}
\label{fig:p2_11}
\end{subfigure}
~
\begin{subfigure}[b]{0.2\textwidth}
\includegraphics[width=\textwidth]{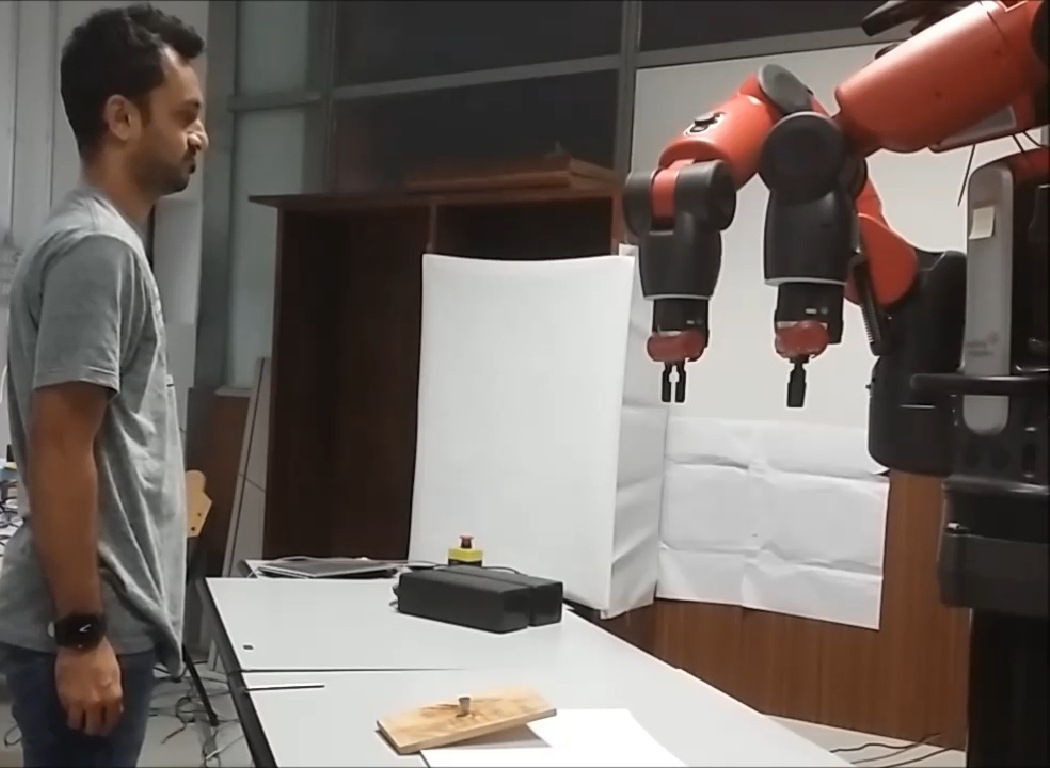}
\caption{}
\label{fig:p2_12}
\end{subfigure}

\caption{The sequence of actions associated with $M_{black}$ after the human decided not to follow $M_{blue}$.}
\label{fig:Path2}
\end{figure*}

As an example of a cooperation model, Figure \ref{fig:Path2} shows snapshots of an actual human-robot cooperation process for the screwing task described above, which starts with $M_{blue}$, but switches to $M_{black}$ when the human performs the first action, which is different from the one expected in $M_{blue}$.
For this task, modeled human and robot actions are: 
\begin{itemize}
\item \textit{initial bolt sink}: the human sinks the bolt in the wooden plate's countersink while the plate is still located on the table (Figures \ref{fig:Path2}a to \ref{fig:Path2}c);
\item \textit{wooden plate pick up and positioning}: the robot picks the wooden plate from a predefined position on the table and keeps it firmly using both grippers (Figure \ref{fig:Path2}d and Figure \ref{fig:Path2}e);
\item \textit{bolt or screwdriver pick up}: the human picks up a bolt or a screwdriver from the table (Figure \ref{fig:Path2}f);
\item \textit{bolt screw}: the human sinks the bolt using the screwdriver (Figure \ref{fig:Path2}g);
\item \textit{screwdriver put down}: the human puts the screwdriver down on the table (Figure \ref{fig:Path2}h and Figure \ref{fig:Path2}i);
\item \textit{wooden plate put down}: the robot puts the wooden plate down in a predefined position on the table (Figure \ref{fig:Path2}j and Figure \ref{fig:Path2}k);
\item \textit{reset pose}: robot's pose is reset (Figure \ref{fig:Path2}l).
\end{itemize}  

In all experiments, each test is considered \textit{successful} if the human and the robot reach the root state of the AND/OR graph, namely \textit{Screwed plate in final position}, by means of any allowed paths.
The human and the robot are allowed to repeat the sequence of actions in a hyper-arc three times before the test is considered failed. 

Experiments have been designed with three specific validation goals in mind:
\begin{itemize}
\item assessing reliability, robustness and flexibility of FlexHRC in terms of cooperation success rate and possible explanations for failures;
\item quantifying computational performance, in terms of latency of action recognition and reasoning time;
\item determining the \textit{Controller} module's capabilities in solving constrained motion problems reactively, i.e., without placing burden on the \textit{Planner} module.
\end{itemize}

\begin{table}[!t]
\caption{Overall information for the tests: number of tests (\#) for each cooperation path $P$, success rate (S), average time (avg) and standard deviation (std) for completing the task successfully.}
\begin{center}
\begin{tabular}{@{}C{1.0cm}C{1.0cm}C{1.2cm}C{1.6cm}C{1.6cm}@{}}
\toprule
$\mathcal{P}$ 			& \# 	& S [\%] 	& avg [sec] 	& std [sec]	\\
\hline
\\
$P_{blue}$ 	& 21 	& 95.24 	& 79.86 		& 3.54 		\\
$P_{black}$	& 23 	& 69.57		& 115.48 		& 23.63		\\
$P_{red}$	& 22 	& 81.82		& 103.50 		& 22.58		\\
\bottomrule
\end{tabular}
\end{center}
\label{tab:SuccFail}
\end{table}
\subsection{Reliability, Robustness and Failures}
A total of $66$ human-robot cooperation experiments have been conducted by a single person.
Table \ref{tab:SuccFail} shows, for each cooperation path $P_{blue}$, $P_{black}$ and $P_{red}$, the number of tests, the success rate, the average completion time and the standard deviation for successful cases. 
It is possible to observe that not all cooperation models are equally difficult. 
$P_{blue}$ is characterized by a very high success rate, whereas the same does not occur for $P_{black}$ and $P_{red}$. 
Both of them are characterized by a greater number of actions, which is reflected by a higher average time needed to complete the operations.
Furthermore, in order for those cooperation models to be successful, more trials of the same action are needed on average (i.e., the human must repeat an operation that has not been detected properly by the \textit{Human Action Recognition} module), and this is reflected by higher values for average standard deviations.
Overall, there are $12$ failures over $66$ experiments, and in particular one failure for $P_{blue}$, seven failures for $P_{black}$ and four failures for $P_{red}$. 
As far as failures are concerned, two of them are due to human mistakes, e.g., misinterpretation of \textit{Planner}'s suggestions, two of them are related to communication failures and temporary high latencies among software modules, one to a robot failure while executing a command, whereas seven of them have been caused by the inability in correctly recognizing human gestures.

A \textit{trend} analysis on all experiments highlights a phenomenon related to how humans adapt their motions in order to facilitate gesture recognition over time.
This means that inertial measurements become more correlated with gesture models encoded using GMM and GMR as the human progresses in performing them.
The most direct consequence is that success rate increases, whereas average completion time and standard deviations decrease. 
In fact, if one looks only at the last $10$ experiments per cooperation path, the success rate for $P_{black}$ and $P_{red}$ become $90\%$ and $93\%$, respectively.
Although adaptation can be expected, current work is devoted to better characterize this phenomenon.
As a preliminary analysis, it can be noticed that it occurs especially for \textit{initial bolt sink}. 
If one looks at this model, it can be observed that it shares similarity with other models to a high degree, such as \textit{bolt or screwdriver pick up} or \textit{screwdriver put down}, mainly in the first part.
After some trials, humans are able to modify their motions to allow \textit{Human Activity Recognition} to disambiguate between all models, with an average increase in successful recognition of roughly $33\%$. 
 
As far as robustness considerations are concerned, the system proves to switch seamlessly among different cooperation models.
Examples of such switches can be observed in the accompanying video\footnote{\url{https://youtu.be/MZv4fUuklq8}}.
Here, let us focus on the switch occurring between $P_{blue}$ and $P_{black}$ in Figures \ref{fig:Path2}a to \ref{fig:Path2}c.
At the beginning of the cooperation, Baxter follows the optimal cooperation path, namely $P_{blue}$, by default.
It moves its right arm to perform \textit{wooden plate pick up and positioning} (Figure \ref{fig:Path2}b) to reach the state \textit{Plate in screwing position}.
However, at the same time, the human decides to perform \textit{initial bolt sink}, i.e., the state \textit{Plate, screw in initial position} is reached.
This state is part of cooperation model $M_{black}$, and therefore FlexHRC sets $P_{black}$ as the current context. 
At this point, the best solution involves reaching \textit{Plate, screw in screwing position}, which means for the robot to perform \textit{wooden plate pick up and positioning}. 
Then, $M_{black}$ unfolds from this moment on.
It is noteworthy that, from the human perspective, this model switch does not imply any perceivable interruption in the operation workflow.
This capability demonstrates requirement $R_1$ described in the Introduction.

\subsection{Computational Performance}

Table \ref{tab:SuccFail} shows in the last two columns the average time required to complete cooperation models and the associated standard deviations.

Overall, $P_{blue}$ outperforms $P_{black}$ and $P_{red}$ in both indicators. 
It is the best one in terms of required time and determinism in execution.
As anticipated above, a lower standard deviation can be explained by the fact that in $P_{black}$ and $P_{red}$ more (human and robot) actions are involved.
Whenever human actions are not recognized by the \textit{Human Action Recognition} module, repetitions are necessary and the time to complete the cooperation increases.
Furthermore, since the likelihood for a human action not to be recognized can be assumed to be normally distributed and unbiased (although adaptation has been observed), since $P_{blue}$ requires less human actions, less false negatives occur and overall execution time is \textit{more} deterministic, on average. 

\begin{table}[!t]
\caption{Reasoning, human, and robot average time percentages for successful tasks.}
\begin{center}
\begin{tabular}{@{}C{1.4cm}C{2.1cm}C{1.9cm}C{1.7cm}@{}}
\toprule
$\mathcal{P}$			& avg $T_{ao}$ [\%]	& avg $T_h$ [\%]	& avg $T_r$ [\%] \\
\hline
\\
$P_{blue}$ 	& 0.09 					& 43.94				& 55.74 \\
$P_{black}$ 	& 0.07 					& 54.50				& 45.29 \\
$P_{red}$ 	& 0.08 					& 56.38				& 43.37 \\
\bottomrule
\end{tabular}
\end{center}
\label{tab:AveTime}
\end{table}
Table \ref{tab:AveTime} reports, for each successfully executed cooperation path, the percentages of average time related to FlexHRC reasoning, and the time needed for human and robot actions.
The first percentage is a measure of the time needed by the AND/OR graph traversal algorithm to suggest next actions to be performed either by the human or the robot. 
Assuming a sequence of $n$ actions, $T_{ao}$ is defined as the sum of all such $n-1$ contributions (the first being set by default on the optimal path), and each contribution is given by the difference between the time $T_{next}$ when the next action $a_i$ suggestion is ready and the time $T_{ack}$ when an acknowledge for a previous action $a_{i-1}$ is received, such that:   
\begin{equation}
T_{ao} = \sum_{i = 2}^{n} T_{next}(a_i) - T_{ack}(a_{i-1}).
\end{equation}
The second percentage refers to the time spent by humans in the cooperation, as well as the time required to detect their motion.
It is noteworthy that this is a \textit{greedy} estimate of human motions, since they may perform other motions not important for the cooperation.   
Let us assume that the human performs $m$ actions, and let us define $\overline{T}_h$ as the sum of all such $m$ contributions, then each contribution depends on the sum of effective human motion time and the time needed by the \textit{Human Action Recognition} module to recognize such a motion.
Therefore, $\overline{T}_h$ is given by:
\begin{equation}
\overline{T}_{{h}} = \sum_{i = 1}^{m} T_{rec}(a_i) - T_{next}(a_{i}), 
\end{equation}
where $T_{rec}$ is the recognition instant.
However, it is also necessary to take into account cooperation model switches.
Assuming to have $k$ context switches during a single cooperation task, an additional term to $\overline{T}_{{h}}$ must be added, which considers the interval between the time $T_{start}$ when the switching action $a_i$ starts and the time $T_{next}$ when the next action $a_{i+1}$ in the new cooperation path is suggested, such as:
\begin{equation}
T_{h} = \overline{T}_{{h}} + \sum_{i = 1}^{k} T_{next}(a_{i+1}) - T_{start}(a_i).
\end{equation}
The average time percentage of AND/OR graph traversing is in all cases less than 0.1\% of the total time. 
As far as $T_h$ and $T_r$ are concerned, it is possible to observe that they are comparable, with the contributions of humans more evident in $P_{black}$ and $P_{red}$. 
This is also due to the proposed control framework, which does not require a computationally intensive planning in the configuration space, thanks to its ability of reactively solving \textit{local} obstacle avoidance constraints.
 
\begin{figure}[!t]
\centering
\includegraphics[width=\columnwidth]{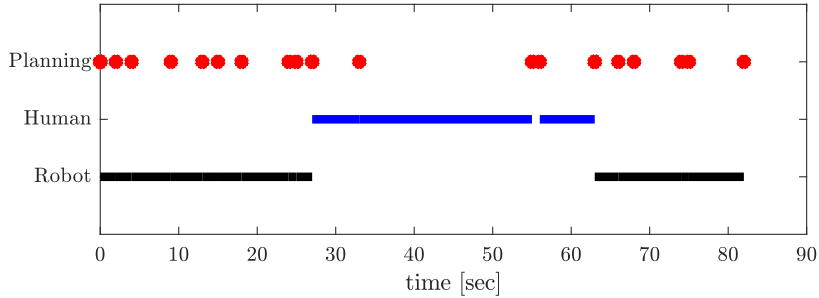}
\caption{An example of time allocation in case of $P_{blue}$.}
\label{fig:HriTiming1}
\end{figure}
Figure \ref{fig:HriTiming1} shows an example of time distribution for one task which follows $P_{blue}$.
The Figure shows that the majority of the time is related to either human or robot actions, and just a negligible part of the whole cooperation task is related to AND/OR reasoning.
The total time to perform the assembly in this test is $82$~s, of which $44.13\%$ spent by the human, $55.56\%$ spent by the robot, and 0.09\% by the \textit{Planner} module.

%

A specific analysis of the delay introduced by the \textit{Human Action Recognition} module during cooperation tasks has also been performed.
In particular, it is necessary to characterize the interval between the time $T_{rec}$ the action $a_i$ is recognized by the module and the time $T_{end}$ the action truly ends.

   
\begin{figure}[!t]
\centering
\begin{subfigure}[b]{1\columnwidth}
\includegraphics[width=\columnwidth]{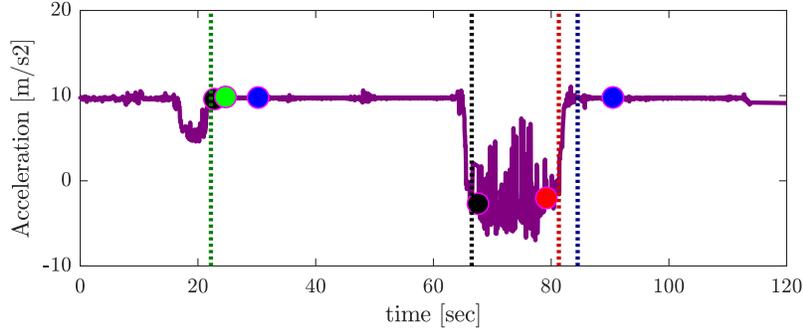}
\caption{Inertial data along $x$ axis.}
\label{fig:possAcc}
\end{subfigure}
\begin{subfigure}[b]{1\columnwidth}
\includegraphics[width=\columnwidth]{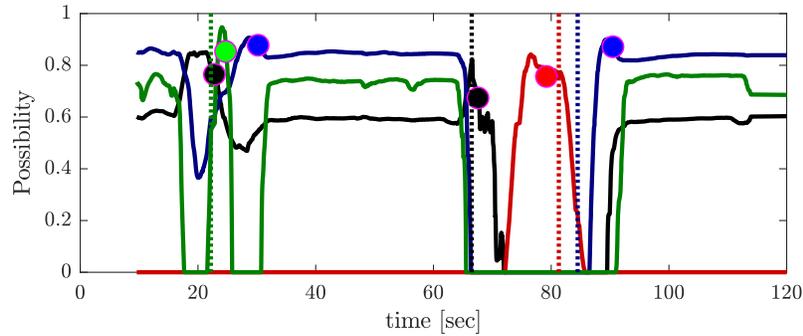}
\caption{Trend in possibilities.}
\label{fig:possPoss}
\end{subfigure}
\caption{Characterization of delays introduced in action recognition for one experiment. Colored dots represent recognition times, and vertical dotted lines are when actual human motions end.}
\label{fig:Delay}
\end{figure}  
Figure \ref{fig:Delay} shows an example of $P_{black}$.
On the top, acceleration data along the $x$ axis are presented, and on the bottom the corresponding possibility trends are shown.
Human actions are represented using different colors (\textit{black} for \textit{bolt or screwdriver pick up}, \textit{red} for \textit{bolt screw}, \textit{blue} for \textit{screwdriver put down}, \textit{green} for \textit{initial bolt sink}).
Colored circles represent the time instants $T_{rec}$ at which \textit{Human Action Recognition} assesses actions, whereas vertical dotted lines show the actual completion instants $T_{end}$, which have been determined by manually inspecting acceleration data. 

In this case, the time delays to recognize \textit{bolt or screwdriver pick up}, \textit{bolt screw}, \textit{screwdriver put down} and \textit{initial bolt sink} are approximately $1.2$~s, $0$~s, $6.1$~s and $2.6$~s. Overall, around $10$~s out of $120$~s of cooperation time are due to the human action recognition delay.
In all the tests, \textit{Human Activity Recognition} introduces a delay roughly less that $10\%$ of the overall cooperation time.




\begin{figure}[!t]
\centering
\begin{subfigure}[b]{0.47\columnwidth}
\includegraphics[width=\columnwidth]{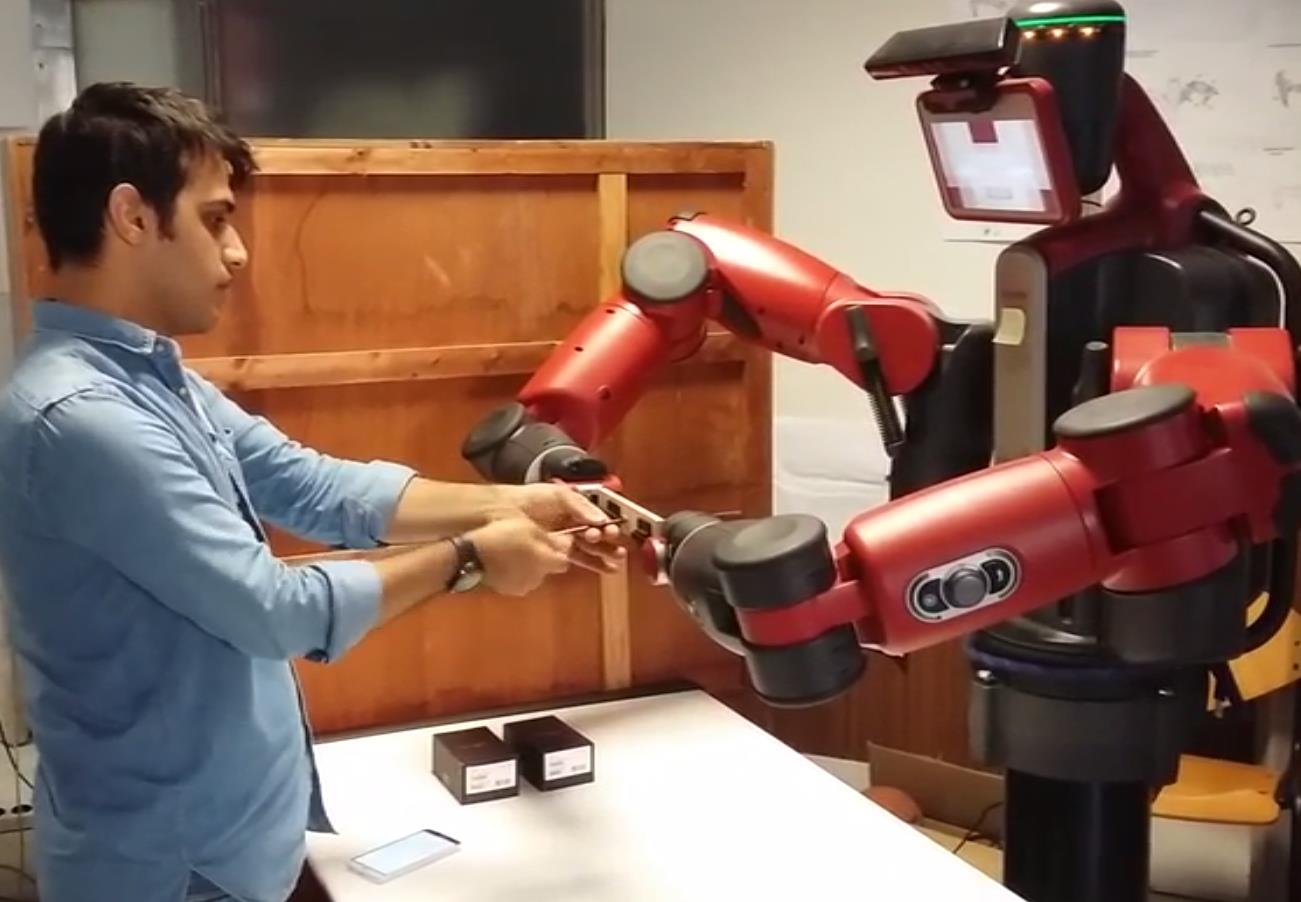}
\caption{}
\label{fig:ObsAvoidTest1a}
\end{subfigure}
\begin{subfigure}[b]{0.47\columnwidth}
\includegraphics[width=\columnwidth]{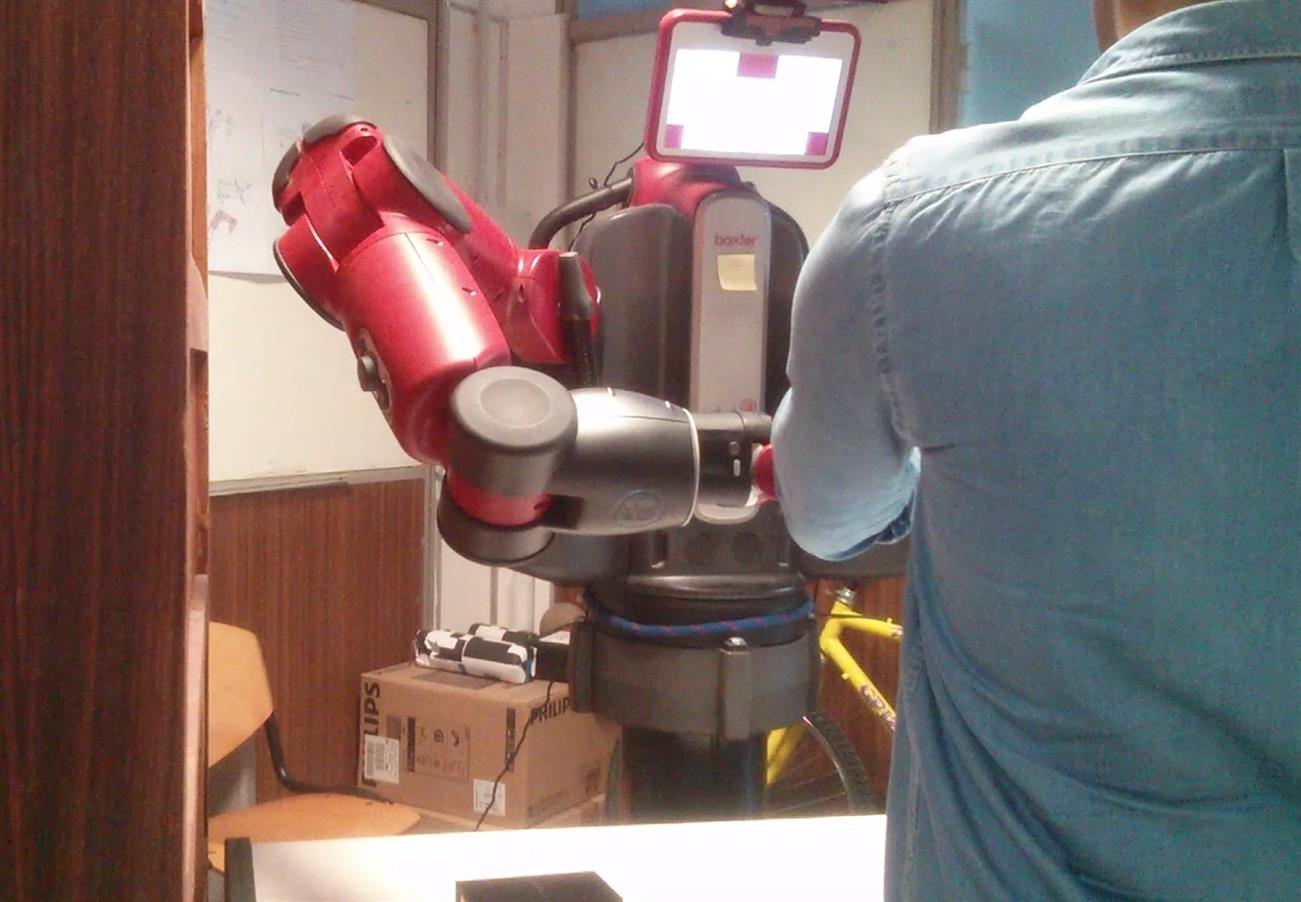}
\caption{}
\label{fig:ObsAvoidTest1b}
\end{subfigure}
\caption{An activity part of $P_{blue}$, when an obstacle is detected and avoided by the elbow joint.}
\label{fig:ObsAvoidTest}
\end{figure}
\begin{figure}[!t]
\includegraphics[width=\columnwidth]{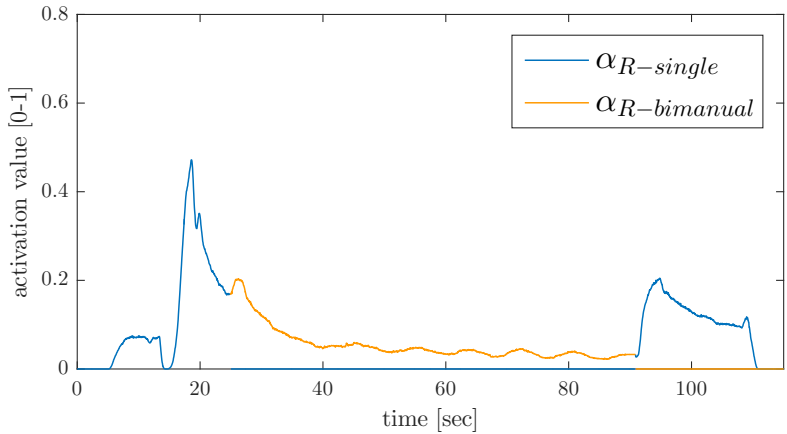}
\caption{Activation function of the elbow avoidance task for the right arm: in \textit{blue} for a single-arm operation, in \textit{orange} for a dual-arm operation.}
\label{fig:exp72_EDAF}	
\end{figure}

\subsection{Task Priority Control}

Tests about the \textit{Controller} module, which embeds the Task Priority control scheme described in Section \ref{sec:taskpriority}, have been carried out.
The metric to test is two-fold: on the one hand, assessing the \textit{Controller} capabilities in dealing with situations requiring reactive control, e.g., obstacle avoidance, which does not need planning in the operational space; on the other hand, doing so without jeopardizing the overall cooperation context process, insofar as cooperation time is concerned. 
A few of the $66$ cooperation tasks have been repeated after the introduction of obstacles in the robot's workspace.

Figure \ref{fig:ObsAvoidTest} shows a cooperation task in which a lateral obstacle has been located on the right hand side of Baxter.
Such an obstacle would impede the robot to perform \textit{wooden plate pick up and positioning}. 
Therefore, beside action-specific control objectives and tasks, safety-oriented tasks have been added, for instance \textit{arm obstacle avoidance}.
Figure \ref{fig:exp72_EDAF} shows the trend of the corresponding activation function for the obstacle avoidance task of \eqref{eq:actfun}, during both single-arm and dual-arm operations.
It is possible to observe that the activation value never reaches the maximum, and the avoidance task is completed within the established thresholds.
Therefore, the \textit{Planner} can focus on the generation of Cartesian trajectories for the end-effectors or the object being manipulated, once grasped by both robot grippers, without planning in the operational space.

As far as the effect of reactive tasks on actual cooperation context execution times, there is no observable difference.
For instance, in this test, avg $T_{ao} = 0.06$, avg $T_h = 54.76$ and avg $T_r = 44.96$. 
Although an exhaustive experimental campaign varying obstacle size and number has not been carried out, these results allow us to conclude that FlexHRC demonstrates requirement $R_2$ as described in the Introduction.

\section{Conclusions}

In this paper, a novel architecture for human-robot cooperation is proposed, which is aimed at addressing a few challenges in shop-floor environments.
FlexHRC supports a natural, intuitive, assisted and direct cooperation.
On the one hand, human gestures implicitly drive the cooperation by executing meaningful actions, and the robot flexibly and seamlessly adapts to those actions via a number of allowed cooperation models.
On the other hand, the robot controller deals with all low-level complexities, e.g., to perform obstacle avoidance in a full reactive fashion, without the need for re-planning in most cases.
FlexHRC has obviously a number of limitations, which are considered as challenges in current research activities. 
\begin{enumerate}
\item The first is that inertial data models obtained via GMM and GMR can be very similar to each other.
This may lead to false positives, and requires processing as much data as possible before action recognition can occur. 
To solve these ambiguities, work is currently carried out to integrate different sensing modalities, such as RGB-D sensors and wearable suits.
However, an adaptation trend on the human side has been observed: people naturally tend to move in such a way as to maximize the likelihood for their actions to be properly recognized. 
This leads to a possible extension, i.e., to include online learning capabilities to perform co-adaptation, as discussed in \cite{Nikolaidis2014, Fragkiadaki2015, Saveriano2015, Menell2016}.
\item The second is represented by an \textit{a priori} assignment of actions to humans or robots, which is due to different capabilities as far as object manipulation is concerned, in a way maybe similar to what has been done in \cite{Mastrogiovannietal2013}.
An on the fly assignment to the human or the robot reflecting the corresponding capabilities would increase to a great extent the flexibility of the cooperation process.
\end{enumerate}

Finally, it is noteworthy that an important aspect to be addressed is the intersection between Task Priority control, motion planning and execution, and the AND/OR graph.
In particular, detecting when the motion controller is stuck in a local minimum, and the development of motion or whole task re-planning for error recovery constitute an important research topic, as demonstrated by a number of contributions in the field \cite{Srivastavaetal2014, Agrawaletal2016}.

\section*{References}

\bibliography{hri}

\end{document}